\documentclass{article}
\usepackage{ijcai25}

\usepackage{times}
\usepackage{soul}
\usepackage{url}
\usepackage[hidelinks]{hyperref}
\usepackage[utf8]{inputenc}
\usepackage[small]{caption}
\usepackage{graphicx}
\usepackage{amsmath}
\usepackage{amsthm}
\usepackage{booktabs}
\usepackage{algorithm}
\usepackage{algorithmic}
\usepackage[switch]{lineno}
\usepackage{xcolor}
\usepackage{pifont}
\usepackage{amssymb}

\usepackage{multicol}
\usepackage{colortbl}
\usepackage{multirow}
\usepackage{arydshln}   

\usepackage{kotex}

\newcommand{\Fref}[1]{Figure~\ref{#1}}

\newcommand{\Sref}[1]{Section~\ref{#1}}
\newcommand{\Tref}[1]{Table~\ref{#1}}

\newcommand{\ours}{B-RIGHT}

\definecolor{figgreen}{RGB}{137, 194, 118}
\definecolor{figorange}{RGB}{231, 152, 118}

\usepackage{subcaption}

\usepackage[many]{tcolorbox} 
\definecolor{main}{HTML}{000000}    
\newtcolorbox[auto counter, number within=section]{boxB}[2][]{
    boxrule = 1.0pt,
    colframe = main,
    rounded corners,
    arc = 5pt   
    enhanced,
    #1,
    title={#2},
    fonttitle = \small
    
}

\title{B-RIGHT: Benchmark Re-evaluation for Integrity in Generalized \\ Human-Object Interaction Testing}

\author{
Yoojin Jang$^{1}$\footnote{These authors contributed equally to this work.}\and
Junsu Kim$^{1,3}$$^*$\and
Hayeon Kim$^{1}$\and
Eun-ki Lee$^{2}$\and\\
Eun-sol Kim$^{2}$\and
Seungryul Baek$^{1}$\And
Jaejun Yoo$^{1}$\footnote{Corresponding author.}\\
\affiliations
$^1$Ulsan National Institute of Science and Technology (UNIST)\\
$^2$Hangyang University\\
$^3$NAVER AI LAB\\
\emails
\{softjin, jjunsssk, rlagkdus705, srbaek, jaejun.yoo\}@unist.ac.kr,
\{alexleee, eunsolkim\}@hanyang.ac.kr}

\begin{document}
\maketitle

\begin{abstract}
    Human-object interaction (HOI) is an essential problem in artificial intelligence (AI) which aims to understand the visual world that involves complex relationships between humans and objects. However, current benchmarks such as HICO-DET face the following limitations: (1) severe class imbalance and (2) varying number of train and test sets for certain classes. These issues can potentially lead to either inflation or deflation of model performance during evaluation, ultimately undermining the reliability of evaluation scores. In this paper, we propose a systematic approach to develop a new class-balanced dataset, \textbf{B}enchmark \textbf{R}e-evaluation for \textbf{I}ntegrity in \textbf{G}eneralized \textbf{H}uman-object \textbf{I}nteraction \textbf{T}esting (\textbf{\ours}), that addresses these imbalanced problems. \ours~achieves class balance by leveraging balancing algorithm and automated generation-and-filtering processes, ensuring an equal number of instances for each HOI class. Furthermore, we design a balanced zero-shot test set to systematically evaluate models on unseen scenario. Re-evaluating existing models using \ours~reveals substantial the reduction of score variance and changes in performance rankings compared to conventional HICO-DET. Our experiments demonstrate that evaluation under balanced conditions ensure more reliable and fair model comparisons. 
\end{abstract}

\begin{figure}[t!]
\centering
\includegraphics[width=1.\columnwidth]{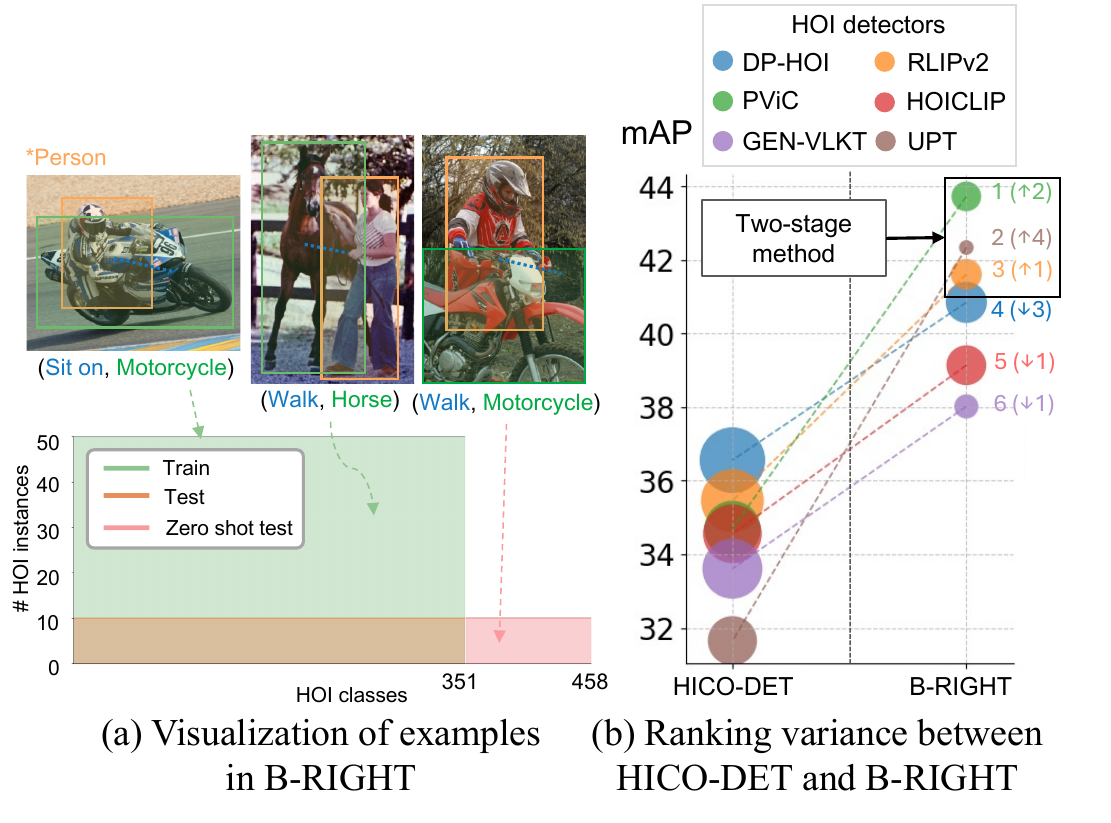}
    \small\caption{(a) Example images of~\ours.
    The proposed dataset ensures a uniform distribution of 351 HOI categories, with 50, 10, and 10 instances for train sets, test sets, and zero-shot evaluation, respectively. (b) Ranking shifts between HICO-DET and~\ours. Circle sizes indicate the variance in class-wise AP scores within each detector, while arrows and numbers denote ranking shifts.}
\label{fig:teaser}
\end{figure}
\section{Introduction}

Human-object interaction (HOI) detection plays a crucial role in enabling artificial intelligence to interpret the visual world, as it captures the complex relationships between humans and objects. Formally, HOI is defined by a triplet (subject, object, interaction), encompassing diverse and subtle actions. 
For example, \Fref{fig:teaser}a illustrates a scenario where ``\texttt{a person is sitting on a motorcycle}," requiring an HOI detector not only to localize the ``\texttt{person}" and the ``\texttt{motorcycle}" but also to classify interactions such as ``\texttt{sit on}," ``\texttt{ride}," or ``\texttt{straddle}."
Advancing HOI detection can benefit a wide range of computer vision applications, including image captioning and reasoning~\cite{wang2021high}, action recognition~\cite{pang2020further}, and localization~\cite{jang2023knowing}.

Recent efforts in HOI detection have largely centered on sophisticated models, spanning from CNN-based models~\cite{chao2018hicodet,gkioxari2018detecting} to Transformer-based architectures~\cite{tamura2021qpic,zhang2023pvic,liao2022genvlkt,xie2023CQL,ning2023hoiclip,zhang2022UPT,li2024disentangled,zhang2021CDN}, leading to notable progress in both accuracy and efficiency.

These methods are predominantly evaluated on established benchmarks like HICO-DET~\cite{chao2018hicodet}, Verb-COCO (V-COCO)~\cite{gupta2015vcoco}, which have shaped the field’s trajectory by providing standardized comparison frameworks. Despite their wide adoption, however, there remains a lack of systematic investigation into the benchmarks themselves, particularly regarding how inherent dataset limitations might skew performance assessments and hinder the reliability of model comparisons.

In particular, HICO-DET, one of the most widely used HOI datasets, exhibits significant class imbalance across both the train and test sets. Moreover, for some classes, the test set even exceeds the train set in size, resulting in problematic data splits that inflate or deflate performance metrics in unpredictable ways. Such imbalances can mask a model's true robustness and undermine fair comparisons, ultimately reducing the reliability of evaluation scores and making it difficult to pinpoint effective improvements. 

\begin{figure}[t!]
  \centering
  \includegraphics[width=1.\columnwidth]{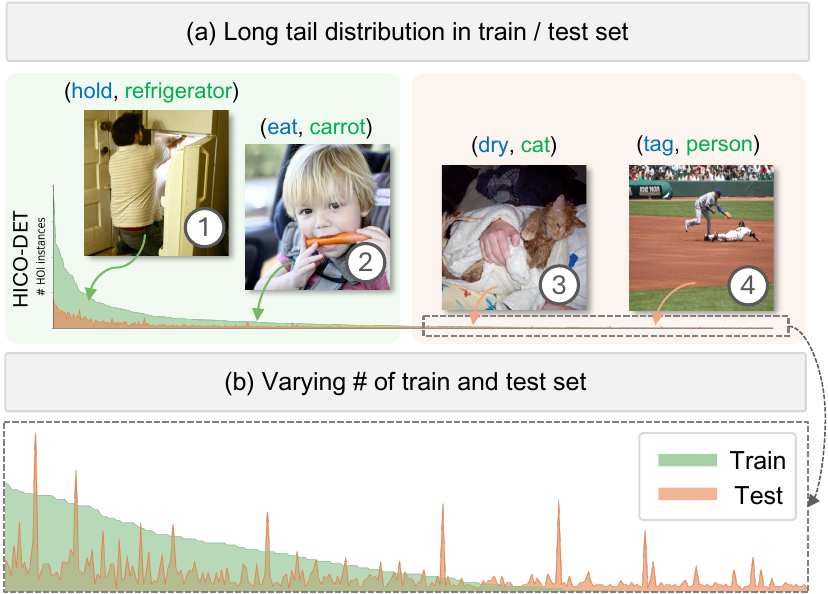}
   \small\caption{\emph{Problem analysis for HOI classes on HICO-DET}: (a) long tail train / test set, (b) varying number of train / test set. The HOI class distribution shows that common classes \textcircled{1} and \textcircled{2} are likely to be well-represented in the real world. However, as we move towards the tail, classes become very rare. Note that for classes in the extreme tail, like \textcircled{3} and  \textcircled{4}, they become particularly rare or ambiguous, highlighting the inherent limitations of HOI detection problem.}
  \label{fig:motivation}
\end{figure}

A seemingly straightforward solution might be to expand the dataset so that each HOI class has an equal number of instances. However, in practice, certain HOI classes (\emph{e.g.}, \texttt{(dry, cat)} or \texttt{(tag, person)}) in \Fref{fig:motivation}a are exceedingly rare or context-dependent and remain underrepresented due to both inherent scarcity and practical constraints such as copyright restrictions on web-crawled samples. Achieving perfect balance in the dataset thus requires substantial time and resources for collection and annotation, an approach that is often considered a very challenging task.

To overcome these challenges in a more practical yet effective manner, we introduce \textbf{B}enchmark \textbf{R}e-evaluation for \textbf{I}ntegrity in \textbf{G}eneralized \textbf{H}uman-object \textbf{I}nteraction \textbf{T}esting (\ours). Rather than attempting to balance every possible HOI class, \ours~carefully selects a diverse subset of classes that can be feasibly balanced. We then augment those classes by combining high-quality synthetic data with web-crawled samples and rigorously filtering out irrelevant or low-fidelity instances using large language models (LLMs) and vision-language models (VLMs). Furthermore, we use LLM and VLM processes to annotate newly collected samples. This automated pipeline ensures that each target class is uniformly represented with sufficient and reliable images. Additionally, we introduce a balanced zero-shot test set for systematically evaluating models on unseen interactions (\Fref{fig:teaser}a). These contributions enable \ours~to set a new benchmark for comprehensive and unbiased HOI evaluation.

In our re-evaluation on the fully class-balanced \ours, we observe a marked reduction in class-wise performance variance (\Fref{fig:teaser}b), indicating that previously inflated or deflated class scores in HICO-DET become more accurately reflected under balanced conditions. Additionally, as illustrated in \Fref{fig:teaser}a, \ours~maintains an equal representation of each HOI category, enabling deeper investigations into how architectural, training, and data-related factors collectively influence model performance. Our comprehensive contributions are organized:

Our main contributions are summarized as follows:
\begin{itemize}
\item We conduct an in-depth analysis of the underexplored aspects of existing HOI benchmark datasets, focusing on the long-tail distribution in train and test sets and various imbalanced ratios.
\item To the best of our knowledge, this is the first work to propose a balanced train and test benchmark, \textbf{\ours}, aimed at evaluating both supervised and zero-shot performance with reliable standards.
\item To construct \textbf{\ours}, we introduce a balancing algorithm and generation-and-filtering schemes to reorganize the dataset and collect high-quality data through generated and crawled images.
\item We retrain and re-evaluate models on \ours, providing detailed analyses of model structures under the balanced dataset.
\end{itemize}

\section{Related works}
\paragraph{HOI benchmarks.}
Various benchmarks for HOI detection have emerged over the years, including V-COCO~\cite{gupta2015vcoco} and HICO-DET~\cite{chao2018hicodet}. While both datasets derive from the MS-COCO dataset, HICO-DET provides more HOI-specific data, involving a broader range of interaction classes and a larger number of images compared to V-COCO. Building upon HICO-DET's broad applicability, recent studies have addressed advanced evaluation scenarios including zero-shot learning~\cite{ning2023hoiclip,yuan2023rlipv2}. However, most of these scenarios have addressed an imbalanced class distribution. In contrast, our work takes the first step toward introducing and evaluating a balanced HOI benchmark dataset, aiming to enable more reliable model evaluations.

\paragraph{HOI methods.}
HOI detection typically comprises two main subtasks: detection and interaction classification. Accordingly, the HOI detectors are categorized into the \emph{one-stage} and \emph{two-stage} architectures. One-stage models integrate detection and interaction classification into a single pipeline, whereas two-stage models first detect objects and then classify the interactions. Recently, most methods incorporate query-based Transformer~\cite{vaswani2017attention} architectures into their schemes, and adopt the DETR family~\cite{carion2020detr,liu2022dab} as a baseline~\cite{tamura2021qpic}.
Further refinements have explored specialized query designs~\cite{zhang2021CDN,xie2023CQL} to better capture interaction classes. Efforts to improve interaction understanding for rare classes also include the utilization of foundation models such as CLIP~\cite{liao2022genvlkt,ning2023hoiclip}, and external knowledge from large-scale datasets~\cite{yuan2023rlipv2,li2024disentangled}.

\section{Motivation}
\label{sec: Motivation}
\paragraph{Potential issues of HOI dataset.}
HICO-DET has become popular due to its diverse categories and scenes. However, it has several limitations that can impair fair and reliable evaluation: As depicted in \Fref{fig:motivation}a,
HICO-DET exhibits an extreme long-tail distribution, with instance counts per class ranging from over 4,000 to just one. This imbalance surpasses that found in common object detection datasets~\cite{lin2014COCO}.
Furthermore, \Fref{fig:motivation}b illustrates the significant variation in the ratio of train to test instances for each class, which is particularly pronounced for rare HOI classes.
As a result, many HOI classes remain substantially underrepresented, skewing model training toward dominant classes and undermining fair comparisons across different approaches.

\begin{figure}[t]
\centering
\includegraphics[width=1.\columnwidth]{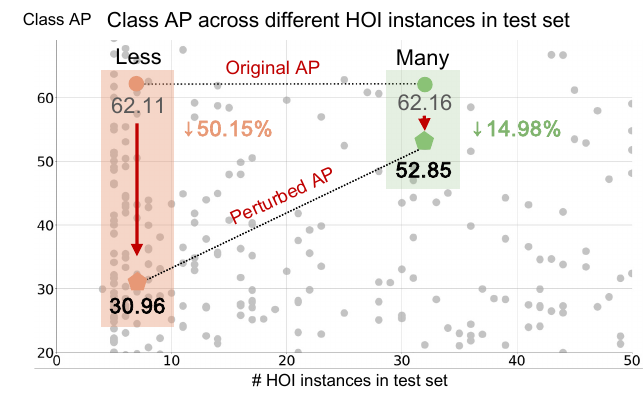}
\small\caption{
    Impact of flipping a single TP instance to FP instance for two classes with similar initial AP scores and train set sizes but different test set sizes. We label the class with fewer test instances as \texttt{Less} (\textcolor{figorange}{orange}) and the class with more instances as \texttt{Many} (\textcolor{figgreen}{green}). 
    Each circle and hexagon represent the original AP and the perturbed AP, respectively, with the numbers below each symbol indicating the AP. The arrows and numbers denote the percentage decrease from the original AP to the perturbed AP.}
\label{fig:data_impact}
\end{figure}

\paragraph{Impact of long-tail dataset on model prediction.} 
It is well known that imbalanced train data can cause overfitting to specific classes~\cite{wang2022chairs} and degrade interaction classification performance~\cite{zhu2023diagnosing,kilickaya2020diagnosing}. 
These biases can be exacerbated when the test set is also imbalanced.
In particular, a smaller number of test instances in certain classes can greatly amplify the impact of each prediction (\emph{i.e}., correct or incorrect) on overall performance metrics. To demonstrate this, we analyze how average precision (AP) changes when a single true positive (TP) is flipped into a false positive (FP), following the method of PViC~\cite{zhang2023pvic}. As depicted in~\Fref{fig:data_impact}, we compare two classes with similar AP scores and training instance counts: \texttt{Many} with more test instances, and \texttt{Less} with fewer. After aligning confidence scores, replacing the highest-confidence TP with an FP results in a much larger reduction in AP for the \texttt{Less} class than for the \texttt{Many} class. This indicates that the number of test data disproportionately affects the final mean average precision (mAP) calculation. 
Consequently, evaluations on imbalanced test sets can distort a model's true performance, particularly for rare classes. This issue leads to high performance variance for these classes, undermining the reliability of the evaluation metrics~\cite{van2024can}. Thus, constructing a balanced test set is crucial for accurately measuring how well models generalize across various human-object interactions.

\section{Balanced dataset construction}
\label{sec: Balanced dataset generation}

The HOI dataset contains multiple classes $\textbf{c}=(c_s, c_v, c_o)\in \mathcal{C}$ in each image $x$, where $\mathcal{C}$ represents the complete set of HOI classes, with $c_s$, $c_v$, and $c_o$ denoting the subject, verb, and object components, respectively. However, simply sampling images from the original HOI dataset is insufficient due to overlapping HOI classes in single images or too few instances per class. In this section, we present the process of constructing a balanced HOI dataset, \ours. To achieve this, we first introduce a \emph{balancing algorithm} to ensure a balanced distribution across all HOI classes for the train and test sets in \Sref{subsec: balancing_algorithm}. We then describe automated \emph{generation-and-filtering schemes} to uniformly augment images in the train set, as detailed in \Sref{subsec: retrieval aug} and \Sref{subsec:filtering_process}.


\begin{figure*}[ht!]
  \centering
  \includegraphics[width=1.\linewidth]{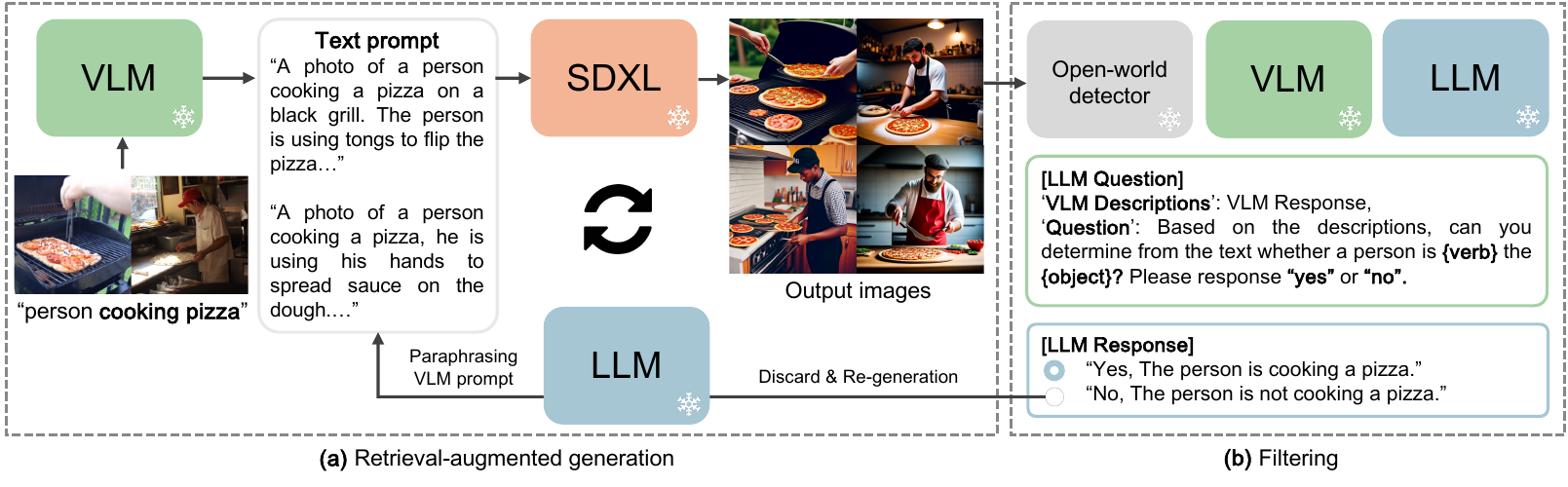} 
  \small\caption{\emph{Overview of our generation-and-filtering schemes}. 
  (a) Retrieval-augmented generation: we retrieve an image from HICO-DET and use a VLM to form a descriptive prompt in a predefined template, which SDXL then uses to create a synthetic image.
  (b) Filtering process: An open-world detector identifies all people and objects, after which another VLM and an LLM verify whether the image correctly depicts the target HOI. Images that do not pass this verification are discarded, and the original prompts are paraphrased to generate new images until we collect enough valid samples or reach our generation limit.}
  \label{fig:RAGfiltering}
\end{figure*}


\subsection{Balancing algorithm}
\label{subsec: balancing_algorithm}
\paragraph{Difficulty of constructing a balanced dataset.}
While random sampling might seem sufficient for creating a balanced dataset, it often fails in practice. For example, a single image may contain multiple HOI pairs. Adding such an image to the dataset based on one HOI can inadvertently increase the counts of other HOI classes present in the same image. This unintended overlap makes it challenging to construct a truly class-balanced dataset. Considering these issues, we propose a balancing algorithm that performs under-sampling with awareness of HOI class distributions.

\paragraph{Balancing algorithm with top-$K$ classes.}
Given a train $\mathcal{D}_{train}$ and test $\mathcal{D}_{test}$ sets from the dataset, we construct a unified dataset $\mathcal{D}_{total}$ by combining $\mathcal{D}_{train}$ and $\mathcal{D}_{test}$ to alleviate the skewed representation of the initial train and test sets. Then, HOI classes, $\mathcal{C}$, in $\mathcal{D}_{total}$ are sorted in descending order based on the number of instances per class. 
Using these sorted classes, $\mathcal{C}_{\text{sorted}}$, we define $L$ as the appropriate number of instances per class that each HOI class should have to be considered balanced. However, it is challenging to augment data to meet both $L_{train}$ and $L_{test}$, especially for tail classes with inherently small numbers of instances. To address this, we select classes with sufficient instances to achieve balance by choosing the top-$K$ classes from the sorted classes $\mathcal{C}_{\text{sorted}}$, denoted as $\mathcal{C}_{\text{sorted}}^{K}$. 
For these top-$K$ classes, we sample images following the balancing algorithm described below:

 1. Starting from the tail classes of $\mathcal{C}_{\text{sorted}}^{K}$ to the head classes, we randomly sample images containing the given class until the instance count reaches at least $L$. We begin by sampling tail classes first to avoid oversampling head classes. 
 
 2. Conversely, from the head classes to the tail classes of $\mathcal{C}_{\text{sorted}}^{K}$, images containing the given class are randomly removed until the instance count is at most $L$.
 
 3. After repeating the above process $N$ times, for any class exceeding $L$ instances, excess instances are randomly removed to match the target count $L$.


\paragraph{Constructing balanced train and test set.} 
Initially, we construct a balanced test set $\mathcal{D}^\text{bal}_{test}$ from the unified total dataset $\mathcal{D}_{total}$ and then form a balanced train set $\mathcal{D}^\text{bal}_{train}$ from the remaining data. The reason for creating the test set first is to ensure that the test split is constituted with the balanced real images; while we allow the train set to have the synthetic images. Since the original data follows a severe long-tail distribution, there may still be classes in the train set with fewer than $L_{train}$ instances even after the balancing process. Additional data collection is necessary for these underrepresented classes, which are typically concentrated toward the lower end of the top-$K$. Within the defined top-$K$ range, we leverage a generative model to produce sufficient data through advanced prompt construction, as detailed in the \Sref{subsec: retrieval aug}.

\subsection{Retrieval-augmented generation}
\label{subsec: retrieval aug}
Recent text-to-image diffusion models (\emph{e.g.}, stable diffusion~\cite{rombach2022high}) have shown remarkable capabilities in generating diverse and high-fidelity images. Leveraging these advances for HOI tasks~\cite{fang2023vil,yang2024mphoi} poses a challenge: standard prompts often fail to capture the nuanced relationships among humans, objects, and their context. To address this, we introduce retrieval-augmented generation approach to refine prompts with rich contextual information, coupled with a multi-step filtering and pseudo-labeling. These methods enable us to augment \ours~with balanced, high-quality synthetic data.

\paragraph{Prompt construction.} 
We adopt a retrieval-augmented approach for constructing prompts to generate high-fidelity images. Given a HOI class $\textbf{c}$, we retrieve a real image $x$ from a subset $\mathcal{X}_\text{\textbf{c}}$ of HICO-DET, which contains images annotated with the HOI class $\textbf{c}$. We then leverage a vision-language model (VLM)~\cite{dong2024internlm2} to produce context-rich textual descriptions of the reference image. To control how the VLM interprets the image, we employ prompt-tuning~\cite{li2021prefixprompt}, as shown following:

\begin{boxB}[label=vlm_prompts]{VLM query.}
\small 
    $<$Image$>$ Please provide a detailed description of the image, focusing on the main person who is \{verb\} a \{obj\}. \ldots 
    Follow this template for your answer: `A photo of a person \{verb\} a/an \{obj\}, \{description\}.'
\end{boxB}
In this query, $<$Image$>$ is a placeholder containing the reference image's encoded features, enabling the VLM to interpret the visual context. The tokens \{verb\} and \{obj\} specify the target interaction and the corresponding object, respectively, while \{description\} captures additional attributes and background details. Once the VLM outputs its descriptive text $p_\text{VLM}$ in a predefined template, we use it as input for text-to-image models.

\paragraph{Image generation.}
\label{subsec:image_generation}
After constructing the prompt $p_\text{VLM}$, we feed it to a text-to-image diffusion model, SDXL-lightning~\cite{lin2024sdxllightning} (denoted as $\Phi_\text{SD}$), to synthesize a high-fidelity image $x'$. Formally, we follow:
\begin{align}
    x' = \Phi_\text{SD}(p_\text{VLM}).
\end{align}

\begin{figure}[t!]

  \centering
  \includegraphics[width=\linewidth]{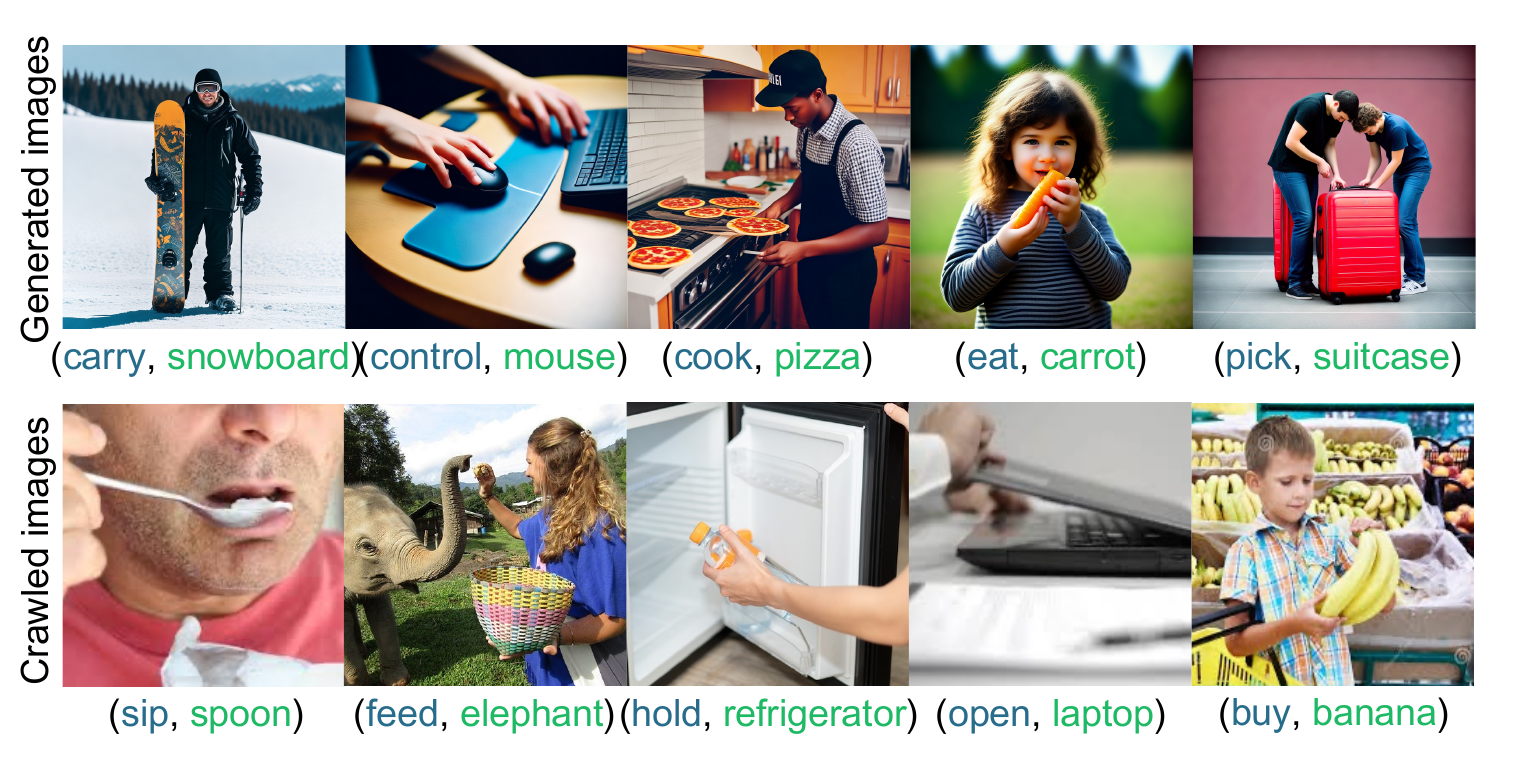}
   \small\caption{Example of the generated and crawled images with \texttt{(verb, object)} pairs after our augmented process.}
  \label{fig:generated_crawled}
\end{figure}

\subsection{Filtering process}
\label{subsec:filtering_process}
Even with careful prompt construction, text-to-image model ($\Phi_\text{SD}$) can produce inaccurate images. To address this, we employ a multi-step filtering strategy that uses both VLM~\cite{dong2024internlm2} and LLM~\cite{jiang2023mistral}. We first use an open-world object detector~\cite{liu2023grounding} to identify all people and objects within each image. For each possible human-object pair, we query a region-based VLM~\cite{you2023ferret} (Section~\ref{subsec: retrieval aug}) to obtain localized descriptions of the bounding-box regions. We then feed these descriptions into an LLM~\cite{jiang2023mistral} using a predefined question prompt, for example:
``\texttt{Based on the description, can you confirm if the person is \{verb\} the \{object\}? Please answer `Yes' or `No'.}"
If the LLM responds with ``Yes," we assume the image accurately depicts the target HOI and retain that sample with final annotations. If the response is ``No," we discard the image or paraphrase the original text prompt via the LLM and regenerate a new image. This process repeats until we obtain enough valid images or exhaust our generation attempts. We summarize the complete generation-and-filtering pipeline in~\Fref{fig:RAGfiltering} and showcase examples of successfully filtered images in~\Fref{fig:generated_crawled}.

Meanwhile, for cases where generated data remain insufficient, we also collect web-crawled images and apply the same filtering process. Additional details regarding data crawling and automated pseudo-labeling methods are provided in the supplementary material.

\begin{table*}[ht!]
\footnotesize
\centering

\begin{tabular}{lccccccc} 
\toprule
\textbf{Dataset} & \textbf{Split} & \textbf{Balanced} & \#\textbf{Images} & \#\textbf{Instances} / \#\textbf{Max} / \#\textbf{Min} & \#\textbf{Classes} & \#\textbf{Object} & \#\textbf{Verb} \\
\midrule
HICO-DET           & Train          & \ding{56} & 38,118           & 117,871 / 4,051 / 1 & 600 & 80  & 117 \\
HICO-DET           & Test           & \ding{56} & 9,658            & 33,405 / 898 / 2    & 600 & 80  & 117 \\
\midrule
B-RIGHT        & Train          & \ding{52} & 6,792            & 17,550 / 50 / 50    & 351 & 78  & 87  \\
B-RIGHT        & Test           & \ding{52} & 1,605            & 3,510 / 10 / 10     & 351 & 78  & 87  \\
B-RIGHT        & ZS-Test        & \ding{52} & 750              & 1,070 / 10 / 10     & 107 & 58  & 40  \\
\bottomrule
\end{tabular}

\small\caption{
Statistics of HICO-DET compared to one of the \ours~ splits. ``\#Instances," ``\#Max," and ``\#Min" represent the total number of instances across all classes, the maximum number of instances within a single class, and the minimum number of instances within a single class, respectively. ``Balanced" indicates whether the dataset split is balanced (\ding{52} = Yes, \ding{56} = No). ``ZS-Test" refers to zero-shot test set. Remaining splits are provided in the supplementary material.}
\label{tab:hicodet_ours_stat}
\end{table*}

\section{Analysis on \ours}
In this section, we introduce our dataset \ours~and discuss how it influences the evaluation of HOI detectors.

\subsection{Experimental settings}
\paragraph{Dataset.} We adopt HICO-DET~\cite{chao2018hicodet} as our baseline dataset, which consists of 38,118 train sets images and 9,658 test sets images across 600 HOI categories (80 objects and 117 verbs). Building on this, we derive \ours, a balanced subset with 351 HOI classes, as detailed in Sections~\ref{sec: Balanced dataset generation}. Each class in \ours~has exactly 50 train sets instances and 10 test sets instances. Further details on the construction process are provided in the supplementary material.
\Tref{tab:hicodet_ours_stat} compares core statistics between HICO-DET and \ours, illustrating how \ours~ eliminates the extreme class imbalances (\emph{e.g.}, from 4,051 vs.\ 1 instance per class in HICO-DET to a fixed 10 in \ours).
Additionally, we construct a balanced zero-shot test set by creating new HOI classes based on novel combinations of objects and verbs that were seen during training but not combined in the 351 selected classes. Unlike prior zero-shot benchmarks~\cite{chao2018hicodet} that rely on imbalanced datasets, the zero-shot split of \ours~ensures a uniform 10 instances per class, facilitating fair evaluation of unseen HOI classes.

\begin{table}[]
\renewcommand{\arraystretch}{1.0}
\footnotesize
\centering
\begin{tabular}{cccc}
\toprule
\textbf{Model name}    &  \textbf{Backbone} & \begin{tabular}[c]{@{}c@{}} \textbf{Pretraining} \\ \textbf{datasets}\end{tabular} & \begin{tabular}[c]{@{}c@{}} \textbf{Auxiliary} \end{tabular} \\
\midrule

\rowcolor{gray!20} \multicolumn{4}{l}{ \textit{One-stage models}} \\\midrule
DP-HOI & DETR & \begin{tabular}[c]{@{}c@{}}\scriptsize HAA500, K700, \\ \scriptsize F30K, VG, \\ \scriptsize COCO, O365\end{tabular} & CLIP \\\hdashline
HOICLIP        & DETR &  \scriptsize COCO & CLIP \\\hdashline
GEN-VLKT       & DETR &  \scriptsize COCO & CLIP \\\hdashline
CDN            & DETR &  \scriptsize COCO & No \\\hdashline
CQL            & DETR &  \scriptsize COCO & No \\\hdashline
QPIC           & DETR &  \scriptsize COCO & No \\
\midrule
\rowcolor{gray!20} \multicolumn{4}{l}{ \textit{Two-stage models}} \\\midrule
RLIPv2      & DAB-DDETR           &  \scriptsize VG, COCO, O365 & No \\\hdashline
PViC        & DETR                &  \scriptsize COCO, HICO-DET$^\dagger$ & No \\\hdashline
UPT         & DETR                &  \scriptsize COCO, HICO-DET$^\dagger$ & No \\\midrule
\bottomrule

\end{tabular}
\caption{Comparison of state-of-the-art HOI detectors evaluated in this study, including their pretraining datasets and auxiliary components. $^\dagger$ indicates HICO-DET is used solely for fine-tuning the DETR backbone, which remains frozen during HOI training.}
\label{tab:model_description}
\end{table}

\paragraph{HOI detectors.} 
\Tref{tab:model_description} summarizes Transformer~\cite{vaswani2017attention} based HOI detectors evaluated in this study. 
We employ a set of one-stage models, including DP-HOI~\cite{li2024disentangled}, HOICLIP~\cite{ning2023hoiclip}, GEN-VLKT~\cite{liao2022genvlkt}, CDN~\cite{zhang2021CDN}, CQL~\cite{xie2023CQL}, and QPIC~\cite{tamura2021qpic}. While most of these rely primarily on COCO for pretraining, DP-HOI incorporates a significantly broader range of datasets, including HAA500~\cite{chung2021haa500}, Kinetics700 (K700)~\cite{carreira2019short}, Flickr30K (F30K)~\cite{young2014image}, Visual Genome (VG)~\cite{krishna2017visual}, and Object365 (O365)~\cite{shao2019objects365}. For two-stage models, we employ RLIPv2~\cite{yuan2023rlipv2}, PViC~\cite{zhang2023pvic}, and UPT~\cite{zhang2022UPT}. They utilize large-scale datasets like VG, COCO, O365, and HICO-DET during pretraining. Most methods adopt DETR~\cite{carion2020detr} architectures, though RLIPv2 uses variants like DAB-Deformable DETR (DAB-DDETR)~\cite{liu2022dab}. 

By including diverse training strategies and architectural designs, we aim to comprehensively evaluate these models under the balanced setting of \ours.

\paragraph{Evaluation metrics.} 
We follow the standard practice of using mean average precision (mAP) to measure HOI detection accuracy. Our evaluation framework consists of two distinct scenarios. In the default scenario, models for HICO-DET are trained and evaluated on all classes, reflecting the dataset's original design. In contrast, for \ours, models are trained and evaluated on 351 classes, which aligns with the dataset's focus on balanced evaluation. Furthermore, we evaluate zero-shot performance scenario using two different splits: HICO-DET's original rare first unseen composition (RF-UC)~\cite{chao2018hicodet}, which exhibits class imbalance, and our newly proposed \ours~zero-shot partition.

\subsection{Analysis of model performance on the HICO-DET and \ours~}
\label{sec:Analysis}

\begin{table*}[th!]
\centering
\footnotesize

\begin{tabular}{lccccccc}
\toprule
\multicolumn{1}{c}{\multirow{2}{*}{}} & \multicolumn{1}{l}{} & \multicolumn{4}{c}{Default}                                                         & \multicolumn{2}{c}{Zero-shot}                  \\ \cmidrule{3-8} 
\multicolumn{1}{c}{}                  & \multicolumn{1}{l}{} & \multicolumn{2}{c}{HICO-DET} & \multicolumn{2}{c}{B-RIGHT}                          & HICO-DET (RF-UC) & B-RIGHT \\ \midrule
\multicolumn{1}{c}{Model}             & Architecture         & mAP           & Rank         & mAP   & Rank                                         & mAP              & mAP        \\ \midrule
DP-HOI                                & one stage            & 36.56         & 1            & 40.85 & 4 \textcolor{blue}{(↓ 3)} & 30.49            & 31.81      \\ \midrule
RLIPv2                                & two stage            & 35.46         & 2            & 41.61 & 3 \textcolor{blue}{(↓ 1)} & 21.45            & 29.97      \\ \midrule
PViC                                  & two stage            & 34.69         & 3            & 43.73 & 1 \textcolor{red}{(↑ 2)}  & --              & 37.07      \\ \midrule
HOICLIP                               & one stage            & 34.56         & 4            & 39.14 & 5 \textcolor{blue}{(↓ 1)} & 25.53            & 28.36      \\ \midrule
GEN-VLKT                              & one stage            & 33.61         & 5            & 38.02 & 6 \textcolor{blue}{(↓ 1)} & 21.36            & 4.38       \\ \midrule
UPT                                   & two stage            & 31.65         & 6            & 42.34 & 2 \textcolor{red}{(↑ 4)}  & --               & 35.95      \\ \midrule
CQL                                   & one stage            & 31.58         & 7            & 33.59 & 9 \textcolor{blue}{(↓ 2)} & --               & 24.68      \\ \midrule
CDN                                   & one stage            & 31.36         & 8            & 35.24 & 7 \textcolor{red}{(↑ 1)}  & --               & 27.05      \\ \midrule
QPIC                                  & one stage            & 29.11         & 9            & 34.00 & 8 \textcolor{red}{(↑ 1)}  & --               & 24.05      \\ \bottomrule
\end{tabular}

\caption{Comparison of rankings of HOI detectors on HICO-DET and \ours. `Rank’ indicates the ranking of models based on their mAP scores, while arrows and numbers denote ranking shifts. The \textcolor{red}{↑} indicates an improvement in ranking compared to HICO-DET, while the \textcolor{blue}{↓} indicates a decrease in ranking. The mAP scores for HICO-DET are referenced from previously reported results in their paper. The symbol (-) indicates models for which official implementation code is unavailable.}

\label{tab:bright_hicodet}
\end{table*}

This section analyzes how state-of-the-art HOI detectors perform when trained on our balanced \ours~dataset compared to HICO-DET, which is imbalanced. We also examine zero-shot performance across both benchmarks.


\begin{figure}[t!]
  \centering
  \includegraphics[width=1.\columnwidth]{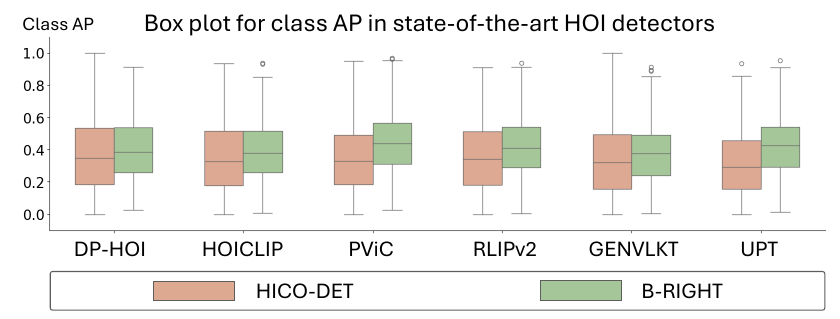}
  \small\caption{Box plots of class AP distributions for various HOI detectors trained and evaluated on HICO-DET (\textcolor{figorange}{orange}) and \ours~(\textcolor{figgreen}{green}). Each box plot shows the median (center line), interquartile range (box), and outliers (dots) for class AP scores in each dataset.}
   
  \label{fig:variance_detector}
\end{figure}

\paragraph{Variance in class-wise AP.}
In \Sref{sec: Motivation}, we highlighted how imbalanced datasets, where test set sample sizes vary significantly across classes, can inflate or deflate AP for certain classes. This  imbalance naturally increases class-wise score variance, which in turn undermines the reliability of metrics. To address this issue, we demonstrate that balanced datasets promote more consistent model performance across all classes, leading to reduced class-wise score variance and enhanced metric reliability. \Fref{fig:variance_detector} visualizes the class AP statistics for various models on both HICO-DET and \ours. For HICO-DET, all models show high class AP variance, indicating inconsistent evaluations across classes. In contrast, \ours~reduces variance and improves overall AP, enabling more uniform class performance. We also observe consistent trends under the zero-shot setting, which is provided in the supplementary material.


\paragraph{Ranking shifts on \ours.} 
\Tref{tab:bright_hicodet} reveals a dramatic change in model rankings when evaluated on our balanced B-RIGHT versus the heavily skewed HICO-DET at the default setting. On HICO-DET, DP-HOI leads with 36.56 mAP, followed by RLIPv2 at 35.46 mAP, while PViC and UPT achieve lower scores of 34.69 and 31.65, respectively. However, on B-RIGHT, certain models exhibit remarkable improvements: PViC leaps to first place at 43.73 mAP, while UPT climbs to second with 42.34 mAP. RLIPv2 maintains a strong performance of 41.61 mAP, but DP-HOI (the former leader) drops to fourth with 40.85 mAP. Other models (\emph{e.g.}, HOICLIP, GEN-VLKT, CQL, CDN, and QPIC) also show increased absolute scores but shifted standings after addressing class imbalance. For instance, CDN improves from eighth to seventh (31.36 to 35.24 mAP), whereas CQL slips from seventh to ninth (31.58 to 33.59 mAP). These findings demonstrate how a long-tail distribution of test set can either mask or amplify a model's strengths, resulting in potentially misleading rankings. Once each HOI class has the same number of instances, the relative performance differences become more apparent, leading to the noticeable shifts shown in \Tref{tab:bright_hicodet}.

\paragraph{Impact of architectural choices.}
Our ranking comparison in \Tref{tab:bright_hicodet} reveals that one-stage models (\emph{e.g.}, DP-HOI, HOICLIP, GEN-VLKT, \emph{etc}) tend to lose their ranking benefit on the balanced \ours~compared to HICO-DET at the default setting. In contrast, two-stage methods (\emph{e.g.}, RLIPv2, PViC, UPT) show significant improvements. Specifically, DP-HOI, which ranks first on HICO-DET, drops to fourth on \ours, while PViC and UPT climb from third and sixth on HICO-DET to first and second on \ours, respectively.

This shift highlights the advantages of two-stage architectures, which decouple object detection from interaction classification. This separation allows them to learn robust, class-balanced representations. Conversely, one-stage models, with their integrated design, are more prone to overfitting frequent interactions in imbalanced datasets and face challenges in generalizing to classes that present low sample counts.
\paragraph{Impact of pretraining datasets.}
DP-HOI and RLIPv2 leverage extensive pretraining datasets, as detailed in the default setting of \Tref{tab:model_description}. Despite this, both models experience significant performance drops on B-RIGHT compared to HICO-DET: DP-HOI falls from first to fourth rank, and RLIPv2 drops from second to third rank, as shown in \Tref{tab:bright_hicodet}. These results indicate that diverse pretraining alone does not ensure improved performance in balanced scenarios. Instead, extensive pretraining may amplify biases toward frequent classes, with balanced benchmarks like B-RIGHT exposing generalization gaps in rare categories. This highlights the importance of aligning pretraining strategies with balanced evaluation settings.

\paragraph{Role of foundation models.}
As shown in \Tref{tab:bright_hicodet}, vision-language foundation models like CLIP~\cite{radford2021clip}, which are integrated into networks such as HOICLIP and GEN-VLKT, align image and text representations to improve semantic understanding. However, these models show consistent ranking decreases on B-RIGHT, indicating that CLIP cannot resolve biases from imbalanced train data.


\paragraph{Backbone architecture and feature representation.} While many HOI detectors adopt a basic DETR backbone, RLIPv2 stands out by integrating a more advanced variant, DAB-DDETR, which typically converges faster and yields richer spatial features. However, despite this sophisticated backbone and its two-stage design, RLIPv2 underperforms compared to other two-stage methods such as PViC and UPT that use standard DETR backbone on \ours~(see \Tref{tab:bright_hicodet}). This suggests that a more powerful backbone alone may not be sufficient under balanced conditions.


\paragraph{Comparison to models in zero-shot balanced test set.}
In \Tref{tab:bright_hicodet}, two-stage architectures (\emph{e.g.}, PViC and UPT) lead the rankings with 37.07 and 35.95 mAP respectively, demonstrating the advantages of separating detection from interaction classification. PViC achieves an mAP of 37.07, with UPT close behind at 35.95. Their explicit two-stage pipeline proves effective at generalizing to unseen verb–object combinations, aligning with our findings in the \emph{default setting}.
Furthermore, DP-HOI and RLIPv2, which utilize specialized zero-shot or large-scale pretraining strategies, achieve 31.81 and 29.97 mAP, respectively. DP-HOI achieves a strong performance of 31.81 mAP, maintaining its competitiveness despite shifting from its previous top ranking in the RF-UC setting. Likewise, RLIPv2 demonstrates resilient performance (29.97 mAP) when transitioning from the RF-UC to the balanced setting, suggesting that its pretraining pipeline is particularly effective at handling balanced zero-shot scenarios.

Notably, HOICLIP and GEN-VLKT, despite both leveraging CLIP-based knowledge for enhanced zero-shot capabilities, achieve lower mAPs of 28.36 and 4.38, respectively. This indicates that while CLIP frameworks are effective in certain zero-shot splits (like the RF-UC setting), they may not consistently perform well across all zero-shot setting, particularly with balanced unseen class distributions. For a more analysis of HOICLIP and GEN-VLKT, please refer to the supplementary materials. 
These findings highlight how zero-shot performance depends on multiple factors, including architectural choices, specific pretraining approaches, and auxiliary network such as CLIP.


\section{Limitations}
Although \ours~alleviates extreme class imbalance, each HOI category is limited to a fixed number of instances, reducing overall coverage compared to larger-scale datasets. In addition, we rely partly on synthetic images and web-crawled samples, which may not capture every nuanced real-world interaction. Nevertheless, this controlled balancing process offers a uniquely unbiased dataset for fair model comparisons, and we believe these limitations can be addressed by future expansions with more diverse data sources and improved generation techniques.


\section{Conclusion}
We addressed the challenge of imbalanced data in HOI detection benchmarks by introducing \ours, a new dataset that guarantees an equal number of instances for each interaction class. This balanced design exposes latent class biases overlooked in previous long-tailed benchmarks, enabling more transparent analysis of architectural innovations, training strategies, and foundational model. Our experiments show that \ours~significantly reduces class-wise variance, revealing new insights such as the consistent advantages of two-stage methods under balanced conditions. By offering a more fair evaluation framework, we hope \ours~will stimulate further progress in HOI detection, allowing researchers to better pinpoint areas for improvement and advance visual reasoning in more robust, equitable ways.

\clearpage
\bibliographystyle{named}
\bibliography{ijcai25}

\clearpage

\section*{\centering B-RIGHT: Benchmark Re-evaluation for Integrity in Generalized \\ Human-Object
Interaction Testing \\ - Supplementary Materials -}

\setcounter{section}{0} 
\renewcommand{\thesection}{\Alph{section}} 

\renewcommand{\thefigure}{\textbf{S}\arabic{figure}}
\setcounter{figure}{0}
\renewcommand{\thetable}{\textbf{S}\arabic{table}}
\setcounter{table}{0}

\section{Dataset access}
\label{supsec:dataset_access}
We provide four versions of the \ours~ train set, test set,  zero-shot test set, and annotation files. These datasets are available for download from the \href{https://github.com/hellog2n/B-RIGHT}{\textcolor{magenta}{LINK}}. Detailed dataset statistics are discussed in \Sref{appsec: dataset statistics}.
https://github.com/hellog2n/B-RIGHT


\section{Dataset details}
\label{appsec: dataset statistics}
\subsection{Generation examples} 
\begin{figure}[ht!]
  \centering
  \includegraphics[width=1\columnwidth]{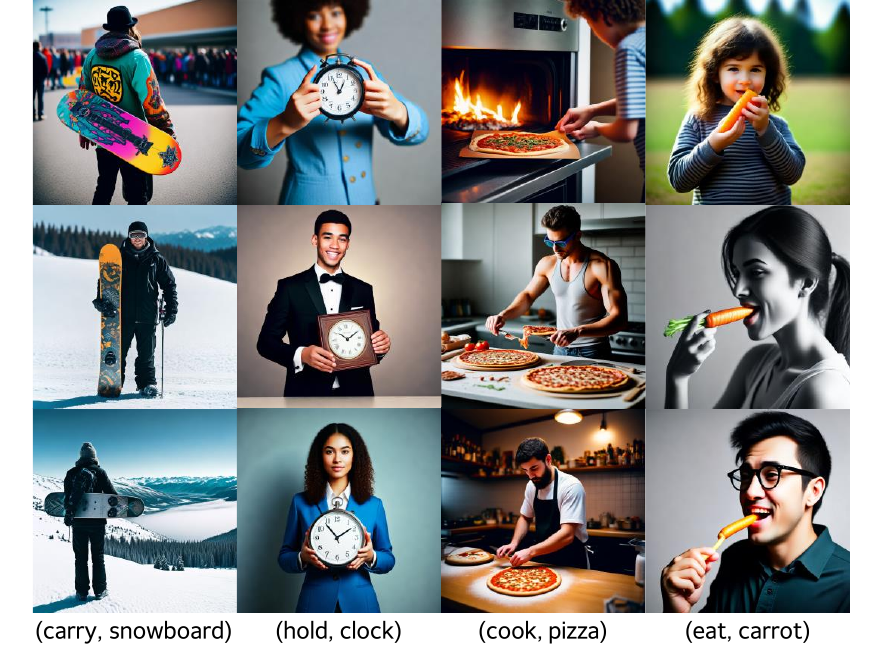}
  \caption{Generated examples in \ours~train set. (verb, object) pairs at the bottom of the figure correspond to HOI class combinations present in HICO-DET.}
  \label{fig:hoi-visualize}
\end{figure}

\autoref{fig:hoi-visualize} displays the image quality remaining after our generation process, using the proposed methods in the main Section~\ref{subsec: retrieval aug}. The generated images represent accurate visual information for HOI and exhibit high quality.

\subsection{VLM prompt examples}
\begin{figure}[ht!]
  \centering
  \includegraphics[width=1\columnwidth]{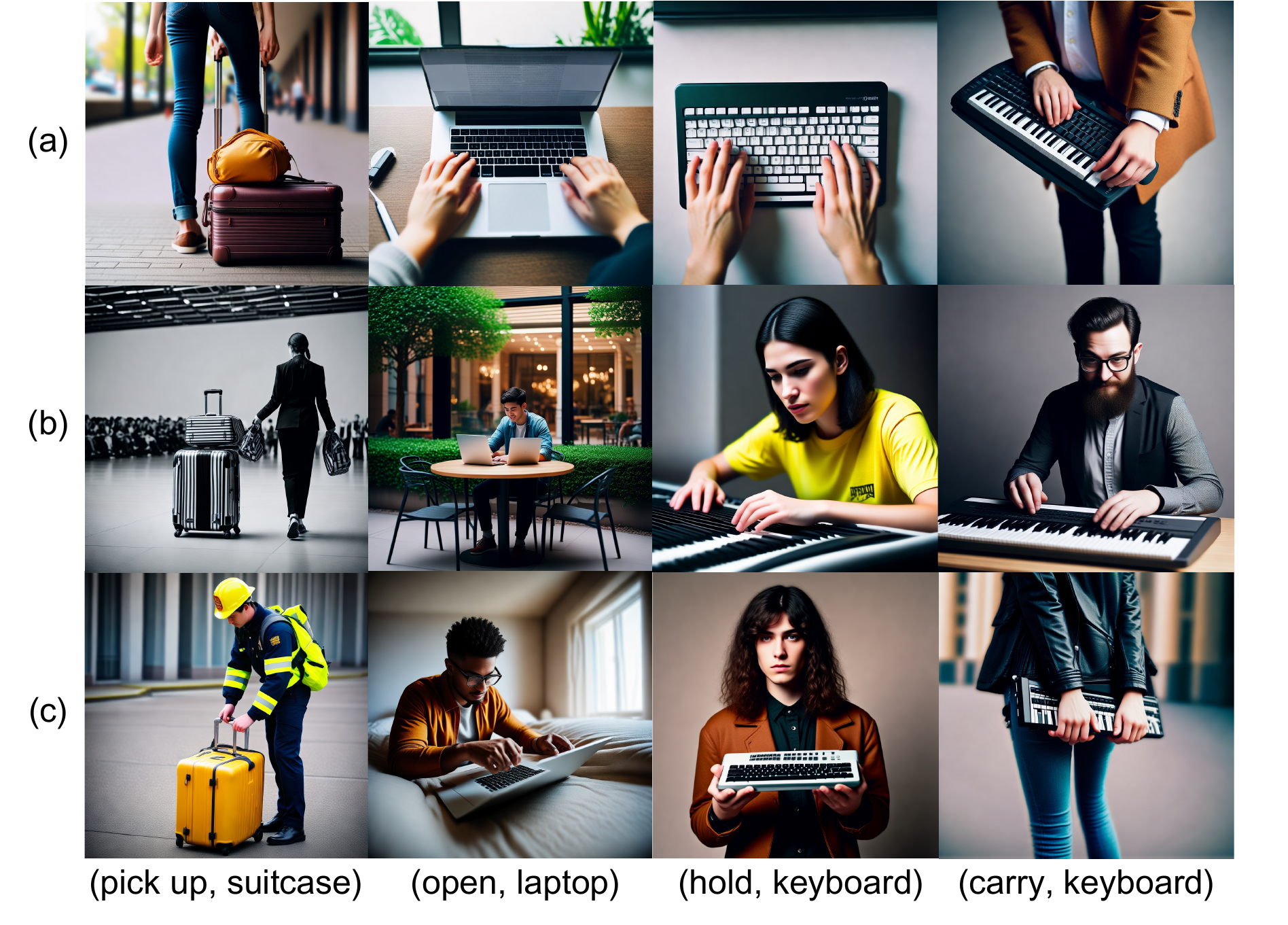}
  \caption{
  Generated examples for various VLM prompts from Stable Diffusion-XL. (a) denotes the basic prompt ``a photo of a person \{verb\}-ing a/an \{object\}", and (b) denotes the basic prompt with long description obtained by the VLM. (c) denotes the basic prompt with long description and processes LLM filtering.}
  \label{appfig:filtering_comparison}
\end{figure}

In~\autoref{appfig:filtering_comparison}, we present the effectiveness of our proposed filtering method (Please see main Section~\ref{subsec:filtering_process}). Compared to (a) basic text-to-image prompts and (b) prompts enhanced with VLM-generated descriptions, (c) our approach combining VLM-enhanced prompts with multi-step filtering consistently produces higher-quality images with more accurate HOIs.

\begin{figure}[ht!]
  \centering
  \includegraphics[width=1\columnwidth]{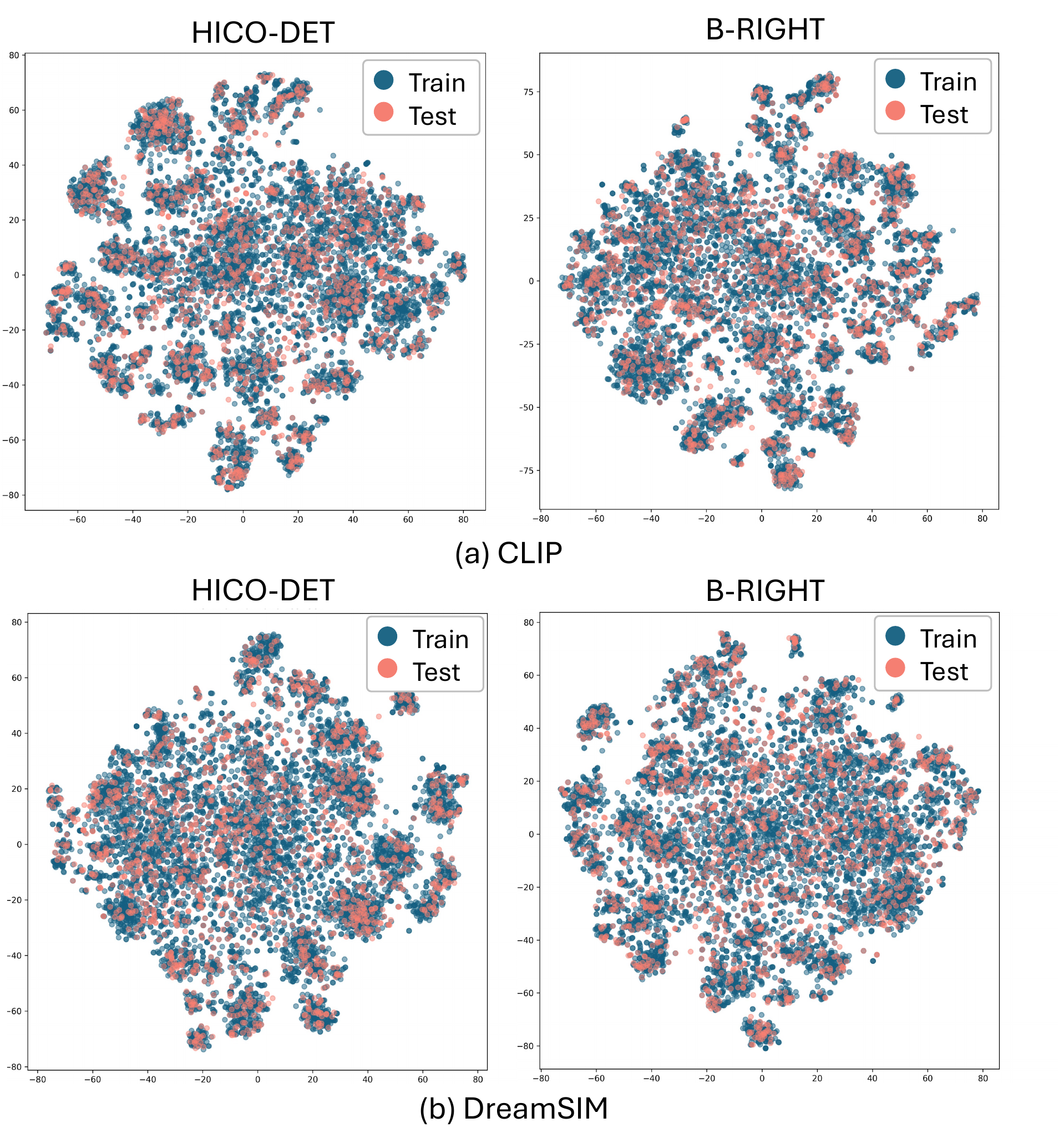}
  \caption{T-SNE visualization for original HICO-DET and our \ours~in (a) CLIP and (b) DreamSIM embedding space. To ensure a fair comparison, we sampled an equal number of images from the original HICO-DET dataset to match the number of images in the \ours~dataset for visualization. Compared to original HICO-DET, our \ours~dataset shows similar distribution between train and test set.}
  \label{fig:distribution of train and testset}
\end{figure}
\subsection{Distribution comparison} 
To compare the similarity between our \ours~dataset and the original HICO-DET~\cite{chao2018hicodet}, we conducted t-SNE visualization, as shown in~\autoref{fig:distribution of train and testset}. For this analysis, we used 6,792 images for the training set and 1,605 images for the test set in \ours. To ensure a fair comparison, we randomly sampled an equal number of images from HICO-DET to match the B-RIGHT dataset. We analyzed the distributions in two embedding spaces: DreamSIM~\cite{fu2023dreamsim}, which measures perceptual visual similarity, and CLIP~\cite{radford2021clip}, which understands visual concepts through language prompts. Our visualization in \autoref{fig:distribution of train and testset} reveals that the clusters and patterns of both distributions are closely match.

\subsection{Dataset statistics} 
\label{appsec: dataset statistics}
To verify our \ours~dataset quantitatively, we present \autoref{apptab: statistics of B-RIGHT}, which provides the statistics of the \ours~train, test and zero-shot test sets. In addition, \autoref{apptab:stat of img in crawled and synthetic} and \autoref{apptab:stat of instance in crawled and synthetic} show the number of images and HOI instances drawn from generated and web-crawled subsets used in the train set, respectively.

\begin{table*}[h!]
\footnotesize
\renewcommand{\arraystretch}{1.2}
\centering

\begin{tabular}{cccccccc}
\toprule
Dataset                                 & Split & \# Images & \# Instances / \#Max / \#Min & \# Categories        & \# Object            & \# Verb  & \# R/A            \\\midrule
\multirow{4}{*}{Train}          & 1     & 6,935     & 17,550 / 50 / 50         & 351                  & 78                   & 87          &  1,171       \\
                                        & 2     & 6,792     & 17,550 / 50 / 50         & 351                  & 78                   & 87     &  1,150            \\
                                        & 3     & 6,900     & 17,550 / 50 / 50         & 351                  & 78                   & 87     &  1,164            \\
                                        & 4     & 6,905     & 17,550 / 50 / 50         & 351                  & 78                   & 87    & 1,197              \\ \midrule
\multirow{4}{*}{Test}           & 1     & 1,618     & 3,510 / 10 / 10          & 351                  & 78                   & 87          &   157      \\
                                        & 2     & 1,605     & 3,510 / 10 / 10          & 351                  & 78                   & 87     &    177          \\
                                        & 3     & 1,579     & 3,510 / 10 / 10          & 351                  & 78                   & 87    &    184           \\
                                        & 4     & 1,594     & 3,510 / 10 / 10          & 351                  & 78                   & 87          &   156      \\
                                        \midrule
\multirow{4}{*}{Zero-shot test} & 1     &  743         & 1,070 / 10 / 10     &    107    &       57    &     38   & 1    \\
  & 2     & 750       & 1,070 / 10 / 10     & 107      & 58     & 40 & 2\\
  & 3     &    751       & 1,070 / 10 / 10     & 107 & 57 & 39 & 2\\
  & 4     &  733         & 1,070 / 10 / 10     &  107 & 58 &38 & 2\\
\bottomrule
\end{tabular}
\caption{Statistics of our various \ours~ train and test sets. \# Instance refers to the total number of HOI instances. \# Max and \# Min denote the maximum and minimum numbers of instances across all HOI categories, respectively. \# R/A represents "Removed Annotation" referring to the count of images from which annotations were removed during the balancing process.}
\label{apptab: statistics of B-RIGHT}
\end{table*}

\begin{table}[h!]
    \footnotesize
    \renewcommand{\arraystretch}{1.2}
    \centering
    \begin{minipage}{0.45\textwidth}
        \centering

        \begin{tabular}{cccc}
            \toprule
            Split & Generated & Crawled & Total \\\midrule
            1 & 56 & 24 & 80 \\
            2 & 57 & 25 & 82 \\
            3 & 63 & 25 & 88 \\
            4 & 59 & 25 & 84 \\
            \bottomrule
        \end{tabular}
        \caption{Statistics of crawled and generated images in train set.}
         \label{apptab:stat of img in crawled and synthetic}
    \end{minipage}%
    \hfill
    \begin{minipage}{0.45\textwidth}
    
        \centering

        \begin{tabular}{cccc}
            \toprule
            Split & Generated & Crawled & Total \\\midrule
            1 & 82 & 39 & 121 \\
            2 & 82 & 38 & 120 \\
            3 & 85 & 40 & 125 \\
            4 & 83 & 39 & 122 \\
            \bottomrule
        \end{tabular}
        \caption{Statistics of HOI instances for crawled and generated images in train set.}
        \label{apptab:stat of instance in crawled and synthetic}
        
    \end{minipage}
\end{table}

\section{Additional balancing process details}
\label{appsec: balancing process details}
\begin{figure*}[h!]
  \centering
  \hspace*{-0.5cm}
  \includegraphics[width=1\linewidth]{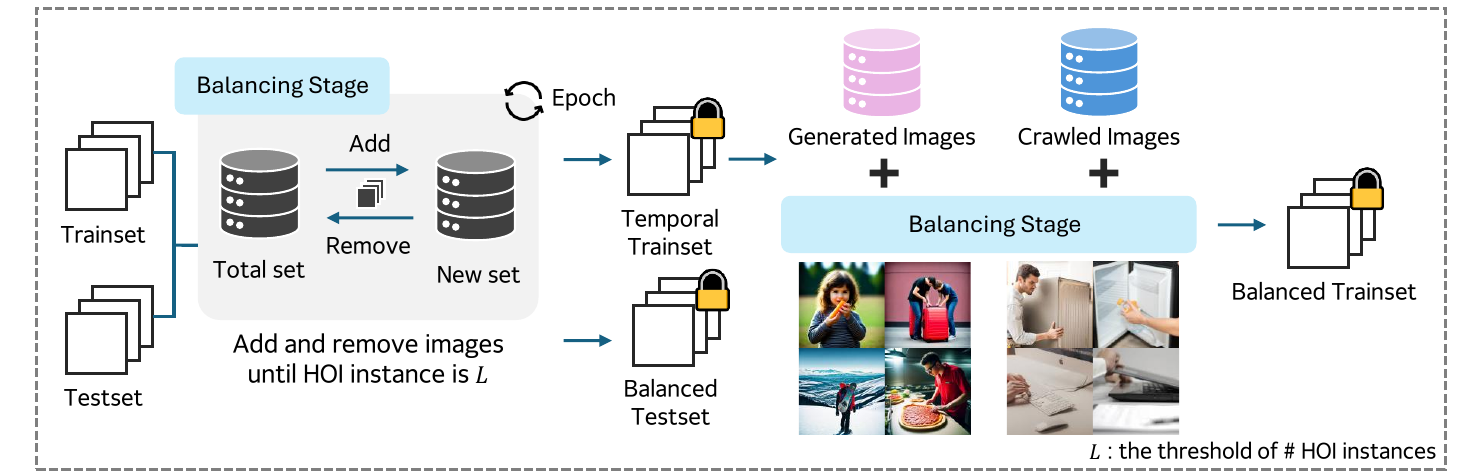}
  \small\caption{An overview of our balancing process. Our balancing process first combines the train and test sets from HICO-DET and then generates new balanced sets. $L$ is the threshold of the number of instances for each HOI class. In this process, by randomly adding and removing images, we ensure that the number of HOI instances for each class exceeds the minimum threshold $L$. Afterward, HOI classes with more than $L$ instances are adjusted to $L$ by randomly deleting the associated annotations. Despite these efforts, some categories may still have fewer than $L$ instances. For these, synthetic or online crawled data are merged to achieve the threshold $L$.}
  
  \label{fig:balance_process}
\end{figure*}

To clarify the balancing procedure described in main Section~\ref{sec: Balanced dataset generation} , \autoref{fig:balance_process} and Algorithm~\ref{Alg:balance_algorithm} provide an overview and pseudo-code, respectively.

\paragraph{Selecting the number of HOI instances $L$.}
For the \ours~test set, we include only real images from HICO-DET. Next, we choose 351 HOI classes that are relatively frequent (\emph{i.e.}, close to the median of around 70 instances across both the HICO-DET train and test sets). Based on this selection, we set the training set to have $L=50$ instances per class and the test set to have $L=10$ instances per class.

\paragraph{Removing annotations for balancing.}
After we finish Algorithm~\ref{Alg:balance_algorithm}, we add extra images if necessary to meet the threshold $L$. Because some classes still exceed $L$, we randomly remove annotations from those classes to ensure a more balanced final distribution.

\begin{algorithm}[tb]
    \caption{Balancing algorithm}
    \label{alg:algorithm}
    \textbf{Input: Total dataset $\mathcal{D}_{total}$, the threshold of instances for each HOI classes $L>0$, Epoch $N=20$. }\\
    \textbf{Define:}
    $ \text{HOI class } \textbf{c} = (c_s, c_v, c_o) \in \mathcal{C}$.   \\
\textbf{Define:} $\mathcal{C}_{total}=\text{get\_hoi\_instances}(\mathcal{D}_{total})$. \\

\textbf{Define:} $  \mathcal{C}^K_{sorted} = \text{sorted\_for\_HOI\_instance}(\mathcal{C}_{total}, K)$    \\
\textbf{Define:} $\mathcal{D}_{bal}=[]$ \hfill

    \textbf{Output}: $\mathcal{D}_{bal}$ 
    \begin{algorithmic}[1] 
        \FOR{$n=1, \dots, N$}
        \STATE // \textcolor{blue}{Adding image stage}
        \FOR{$\textbf{c}=K, \ldots, 1$}
        \WHILE{$ \text{count\_HOI\_instance}(\mathcal{D}_{bal}, \textbf{c}) < L$}
        \STATE $x \leftarrow \text{get\_image}(\textbf{c}, \mathcal{D}_{total}).$ \\

        \IF {$\exists x \in \mathcal{D}_{bal}$}
            \STATE continue\hfill// \textcolor{blue}{If $x$ is in the $\mathcal{D}_{bal}$, pass this image.} \\
        \ENDIF \\
        $\mathcal{D}_{bal} \leftarrow \text{add\_image}(\mathcal{D}_{bal}, x)$.
        \ENDWHILE
        \ENDFOR
        \IF {$ n < N$}
        \STATE // \textcolor{blue}{Removing image stage} \\
        
        \FOR{$\textbf{c}=1, \dots, K$}
            \WHILE{$\text{count\_HOI\_instance}(\mathcal{D}_{bal}, \textbf{c}) > L$}
                \STATE $x \leftarrow \text{get\_image}(\textbf{c}, \mathcal{D}_{total})$. \\
                \IF {$\exists x \in \mathcal{D}_{bal}$}
            \STATE $\mathcal{D}_{bal} \leftarrow \text{remove\_image}(\mathcal{D}_{bal}, x)$. \\
        \ENDIF \\
                
        \ENDWHILE
        \ENDFOR
        \ENDIF
        \ENDFOR
        \STATE \textbf{return} $\mathcal{D}_{bal}$ 
    \end{algorithmic}
\label{Alg:balance_algorithm}
\end{algorithm}

\section{Additional generation process details}
\subsection{Stable Diffusion} 
Stable diffusion (SD)~\cite{rombach2022high} incorporates a VAE~\cite{kingma2013auto} structure to first extract the latent vector $z \in \mathbb{R}^{64 \times 64}$ from the image $x \in \mathbb{R}^{512 \times 512}$ and generate reconstructed images $\hat{x}$ from the latent vector $z$. This process reduces computational costs by operating in a lower-dimensional latent space rather than the original high-dimensional image space. SD also utilizes a U-Net~\cite{ronneberger2015unet} architecture with a varied diffuse algorithms~\cite{ho2020ddpm,song2020ddim} 
framework, where Gaussian noise is added to the latent vector and subsequently denoised during the reverse diffusion process. 

The core function of SD is $f_\theta(z_t, t, \textbf{T})$, where the trained U-Net is employed for $f_\theta$, $t$ denotes the time embedding and $z_t$ represents the latent representation at the $t$-th diffusion time step. Furthermore, SD utilizes the textual information $\textbf{T}$ extracted by CLIP's~\cite{radford2021clip} text encoder. It is then injected via cross-attention mechanisms to efficiently generate images based on the texture information $\textbf{T}$. 

Thus the SD's variants~\cite{stable_diffusion,podell2023sdxl,lin2024sdxllightning} demonstrate remarkable versatility and have recently emerged as a powerful tool for a wide range of tasks, showcasing their adaptability and effectiveness across diverse applications~\cite{fang2023vil,yang2024mphoi,kim2024sddgr}.

\subsection{Vision-Language Models}
\paragraph{Image description models.} 
Vision-language models (VLMs) have significantly evolved, initially focusing on short image captioning tasks and now expanding to complex tasks like question-answering and long descriptions. This progress has been facilitated by integrating large language models (LLMs) with vision components, allowing for more sophisticated image understanding and generation capabilities.  
InternLM2v~\cite{dong2024internlm2} 
leverages LoRA~\cite{hu2021lora} to selectively train visual feature extraction modules, enhancing its ability to generate detailed and contextually rich image captions. 
Honeybee~\cite{cha2023honeybee}, based on the vicuna~\cite{vicuna2023}, utilizes innovative visual projectors, including CNNs~\cite{he2016resnet} and Transformers~\cite{zhu2020deformable}, to interpret and describe visual scenes accurately. 
In our study, we used two models ~\cite{cha2023honeybee,dong2024internlm2} in the main \Sref{subsec:filtering_process} depending on resource availability.

\paragraph{Region-based model.} 
Ferret~\cite{you2023ferret} is an outstanding model in the field of a specific-region VLM, designed to facilitate precise question-answering for specific regions within images. It utilizes a hybrid approach to region representation, combining discrete elements (such as points and boxes) with continuous features (like strokes and complex polygons). Leveraging the CLIP~\cite{radford2021clip} for visual feature extraction, Ferret processes images resized to 336 × 336 pixels using CLIP's image encoder to generate feature embeddings $ Z \in \mathbb{R}^{H \times W \times C}$, where $H$ represents height, $W$ width, and $C$ channels of the image. The textual embeddings are created using a pre-trained language model's~\cite{vicuna2023} tokenizer, resulting in embeddings $\textbf{T} \in \mathbb{R}^{L \times D}$ where $L$ is the sequence length and $D$ is the embedding dimension. The visual embeddings are projected to align with the textual embedding dimension $D$ enabling integrated processing of visual and textual data.

To seamlessly incorporate visual information into text prompts, Ferret uses a placeholder denoted by \texttt{$<$Image$>$} for the features over the entire image. When addressing specific regions within an image, the question is normally formatted as “\texttt{Is this a [label] [location] $<$SPECIAL$>$?},” where \texttt{[label]} corresponds to specific category label in HICO-DET, and \texttt{$<$SPECIAL$>$} placeholders are replaced with features extracted by the CLIP image encoder from the specified \texttt{[location]} coordinates (\emph{i.e.}, formatted as $[x_1, y_1, x_2, y_2]$) within the image. For example, the question is ``\texttt{Is this a banana [10, 10, 100, 100] $<$SPECIAL$>$?}."

Based on the aforementioned versatility, VLMs~\cite{you2023ferret,internlmxcomposer2,liu2024llava} are utilized for various recent works~\cite{cao2024unihoi,zheng2023openvlm,kim2024vlmpl}.

\subsection{Filtering details}
Here, we cover the filtering details from the main Section~\ref{subsec:filtering_process}. To maximize the effeteness of VLM~\cite{cha2023honeybee} for image description, we avoid constructing prompts that focus narrowly on specific interactions. Instead, we formulate our queries to request as detailed an interpretation as possible of the entire image. This approach ensures that the VLM provides comprehensive descriptions, covering various aspects of the image, such as the main subjects, objects, their interactions, and the surrounding context. The specific VLM query is designed as follows:
\begin{boxB}[label=vlm-prompt]{VLM query for detailed image description.}
    \small
    $<$Image$>$ Please provide a detailed description of this image. Specifically, focus on the person, object, interaction with each other, and all background elements.
\end{boxB}
Here, $<$Image$>$ serves as a placeholder where the image features extracted by CLIP~\cite{radford2021clip}. This query aims to extract a thorough understanding of the entire image, ensuring that the VLM considers all relevant details and contexts.

Following this initial VLM analysis, we then employ the LLM to further refine the filtering process. The LLM's advanced capabilities in text summarization, question-answering, and reasoning are leveraged to ensure the accuracy and relevance of the image descriptions provided by the VLM. Specifically, we use the LLM to verify whether the detailed description accurately includes the intended human-object interaction components: the person, the object (\{obj\}), and the action (\{verb\}). For the verification process, we carefully design the query to confirm the presence and clarity of these interaction components:
\begin{boxB}[label=llm-query]{LLM query for the verification process.}
\small
    $<$Image descriptions$>$ This text provides a detailed description of the image. Your task is to determine if there is a Human-Object Interaction based on the questions I'm asking. Can you definitively determine from the text whether a person is performing the action \{verb\} on the object \{obj\}? If either the person or the \{obj\} is not present, or if you cannot clearly determine the \{verb\} action, respond with `no'. Only respond with `yes' if you can definitively determine the action, person, and \{obj\} (including things similar to the \{obj\}, e.g., sunglasses can be included in the glass class) based on the text. Please begin your response with `yes' or `no', followed by your explanation.
\end{boxB}
Here, $<$Image description$>$ represents a placeholder for descriptions created by VLM~\cite{cha2023honeybee}. We filter out the synthetic images based on the LLM's answers in response to the query. Images that receive a ``yes'' response are retained, ensuring they contain the desired interaction and target object. Conversely, images that receive a ``no'' response are discarded to maintain the quality and relevance of the dataset. This filtering strategy, combining the strengths of both VLM and LLM, allows us to refine the synthetic dataset, ensuring high fidelity to the human-object interactions depicted in the images.

\subsection{Automated pseudo-labeling process}
\label{appsec:automated-pseudo-labeling}
Following the filtering process, we conduct pseudo-labeling for pairs of subjects and objects within the image.  To address this, we use the region-based VLM~\cite{you2023ferret} for question-answering about specific region interactions. We initially construct the HOI pair candidates associated with the target objects for each subject. Subsequently, we design prompts for each candidate to query the region-based VLM~\cite{you2023ferret}, incorporating the location information of both the target object and the person to determine if the pairs are correctly identified. Interaction candidates confirmed by the VLM with ``yes" response are considered as the final annotations. The pseudo-labeling procedure is applied to both synthetic and crawled images. An example of the prompt is as follows:
\begin{boxB}[label=ferret-prompt]{VLM query for pseudo labeling.}
\small
    $<$Image$>$ Considering the image, can you definitively determine that \{subject\} $<$region1$>$ is \{verb\} \{object\} $<$region2$>$ in the image? Please respond with `yes' or `no', followed by your explanation.
\end{boxB}
Here, $<$Image$>$ refers to the entire image representation, \{subject\} $<$region1$>$ indicates the region representation of the subject, and \{object\} $<$region2$>$ specifies the region representation of the object.

\section{Human object interaction detector details}

\noindent \textbf{UPT}~\cite{zhang2022UPT} introduces a two-stage Unary-Pairwise Transformer architecture for Human-Object Interactions (HOIs), utilizing both unary and pairwise representations. Initially, UPT identifies and localizes objects, then employs transformer-based object queries to determine interactions between objects and subjects, enhancing classification precision through its sequential detection approach.

\noindent \textbf{HOICLIP}~\cite{ning2023hoiclip} introduces a method for HOI detection that utilizes the pre-trained CLIP model for efficient knowledge transfer. It aligns visual features with corresponding linguistic features through a vision-language model, enhancing the interaction understanding between humans and objects. Moreover, CLIP is used to generate a classifier in a Transformer-based decoder.

\noindent \textbf{CDN}~\cite{zhang2021CDN} introduces a hybrid model that combines a CNN-Transformer for visual feature extraction with two cascade transformer-based decoders. The first decoder identifies human-object bounding-box pairs, while the second predicts action categories for these pairs. This method combines the advantages of both one-stage and two-stage detection methods to improve the accuracy and efficiency of HOI detection, positioning it as a one-stage detector.

\noindent \textbf{QPIC}~\cite{tamura2021qpic} introduces a Transformer-based architecture for Human-Object Interaction (HOI) detection that uses learnable queries to capture interactions between humans and objects. The model aggregates contextual information from the entire image to leverage image-wide context. It also uses an attention mechanism for a better understanding of human-object interactions. Moreover, it constrains queries to detect human-object pairs, preventing duplicate issues within an image.

\noindent \textbf{CQL}~\cite{xie2023CQL} introduces a novel Transformer-based architecture for HOI classification that leverages category-specific queries to enhance detection accuracy. The model employs these queries to concentrate on relevant interactions, gathering contextual information from the entire image. CQL is utilized as an additional component in conjunction with the proposed Transformer-based methods~\cite{tamura2021qpic,liao2022genvlkt}.

\noindent \textbf{RLIPv2}~\cite{yuan2023rlipv2} presents a fast-scaling model for Relational Language-Image Pre-training (RLIP), designed to improve HOI detection. It is pre-trained on large datasets with tagged visual and linguistic information, leveraging a pre-trained relational language-image model like BLIP. The architecture scales efficiently by optimizing the interaction between visual features and textual descriptions, leading to improved performance.

\noindent \textbf{PVIC}~\cite{zhang2023pvic} enhances the detection of HOI by enriching contextual cues for human-object pairs. It integrates fine-grained details and relevant scene contexts through an improved query design and tailored positional embedding, providing a more comprehensive understanding of interactions. Furthermore, PVIC explores an optimized key-value combination of cross-attention. Consequently, it greatly enhances the accuracy and robustness of HOI detection.

\noindent \textbf{GEN-VLKT}~\cite{liao2022genvlkt} introduces Guided-Embedding Network (GEN) and Visual-Linguistic Knowledge Transfer (VLKT) to improve HOI detection. GEN replaces complex post-matching by employing Position Guided Embedding (p-GE) to associate humans and objects as pairs and Instance Guided Embedding (i-GE) to generate interaction queries and classify interactions. VLKT leverages CLIP by initializing the classifier with text embeddings and aligning visual features via mimic loss, addressing long-tailed distribution.

\noindent \textbf{DP-HOI}~\cite{li2024disentangled} introduces a disentangled pre-training framework for HOI detection using object detection and action recognition datasets. The framework consists of an object detection branch, which uses a detection decoder to predict bounding boxes and object classes, and a verb classification branch, which utilizes Reliable Person Queries (RPQs) generated by the detection decoder. These RPQs are passed to a transformer-based interaction decoder to predict verb classes. DP-HOI demonstrates superior performance, particularly in rare and zero-shot HOI classes.
\section{Implementation details}
In this section, we provide a detailed description of dataset collection, training and evaluation settings of the models used in our experiments.
 
\paragraph{Crawled dataset collection.}
To collect the crawled data, we used the search query \textit{`a photo of a/an person \{verb\}-ing a/an \{object\}'} to gather images from internet engines. We then used a pretrained watermark detector\footnote{https://github.com/boomb0om/watermark-detection} to identify and remove images containing watermarks. Following this, the remaining images were filtered using the VLM LLM filtering method described in the main Section~\ref{subsec:filtering_process}. Finally, human filtering and the pseudo-labeling process described in \Sref{appsec:automated-pseudo-labeling} are conducted to ensure high-quality data.

\paragraph{Human filtering process.}
To ensure the high quality of images used for train set, we manually filtered the additional images except for HICO-DET dataset. For this, four individuals conducted the filtering process, remaining only those images that meet both of the following criteria:
\begin{boxB}[label=human-manual]{Manually filtering criteria}
    \small
    (1) Does the image contain both a person and a/an \{object\}? \\
(2) Are the person and a/an \{object\} \{verb\} interacting in the image?
\end{boxB}

\paragraph{Human-object interaction detector configuration.}
We conducted all experiments using 4 A100 80GB GPUs.
We trained all models following the official training code and hyperparameters. For all models, we utilized ResNet-50~\cite{he2016resnet} backbone architecture which is pretrained on MS-COCO~\cite{lin2014COCO}. For RLIPv2, we utilized ResNet-50 as the backbone architecture which was pretrained on COCO~\cite{lin2014COCO}, Visual Genome~\cite{krishna2017visual} and Object365~\cite{shao2019objects365}.
For PViC~\cite{zhang2023pvic} and UPT~\cite{zhang2022UPT}, we utilized ResNet-50 backbone architecture which is finetuned on HICO-DET, following official setting.

\begin{figure}[t!]
  \centering
  \includegraphics[width=1.\columnwidth]{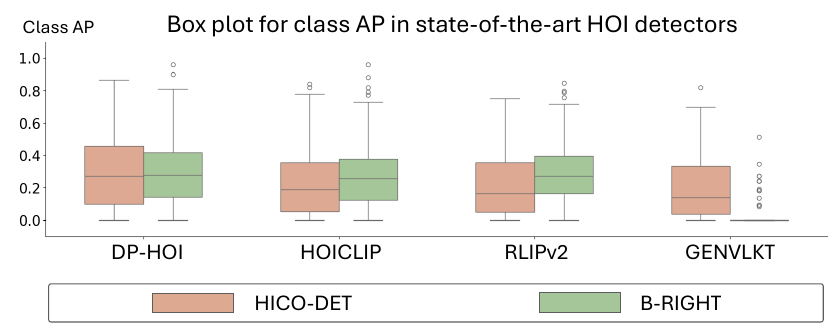}
  \small\caption{Box plots of class AP distributions for various HOI detectors trained and evaluated on HICO-DET RF-UC setting (\textcolor{figorange}{orange}) and \ours~zero shot test (\textcolor{figgreen}{green}). Each box plot shows the median (center line), interquartile range (box), and outliers (dots) for class AP scores in each dataset.}
  \label{fig:zero_shot_variance_detector}
\end{figure}

\begin{figure}[t!]
  \centering
  \includegraphics[width=1.\columnwidth]{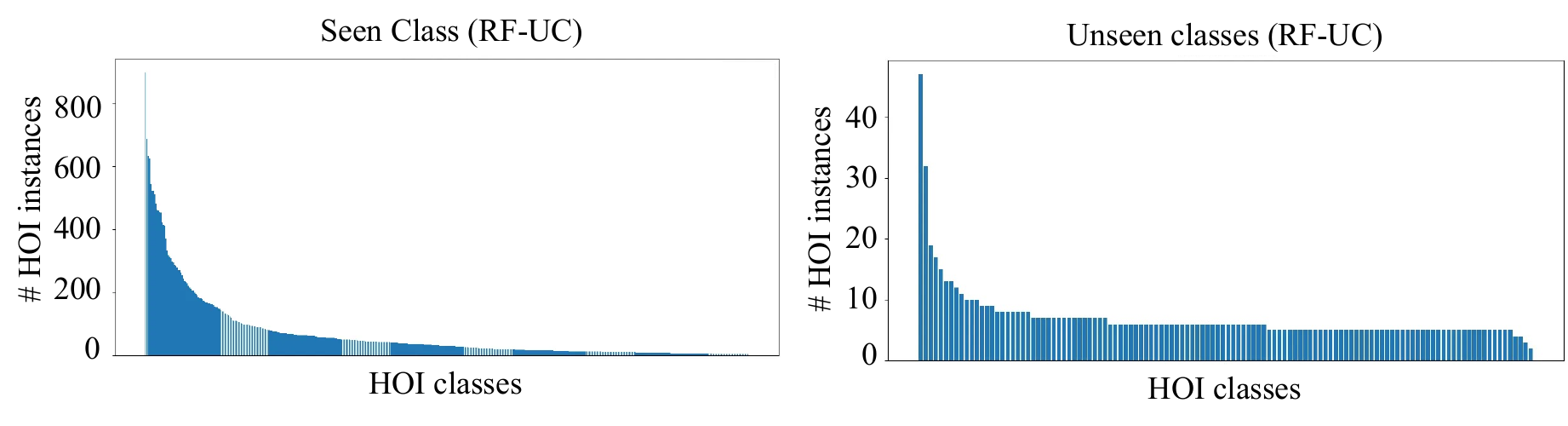}
   \small\caption{Statistics for HOI categories on RF-UC zero-shot test in HICO-DET. HICO-DET zero-shot setting also exhibits imbalanced seen and unseen classes.}
  \label{fig:appen-zeroshotstat}
\end{figure}


\section{Addional anaylsis}
\label{sup:additional_analysis}

\paragraph{Variance in class-wise AP in zero-shot test setting.}
\autoref{fig:zero_shot_variance_detector} illustrates the score variance in class-wise APs of models evaluated under the zero-shot setting discussed in the main \Sref{sec:Analysis}. Similarly, as shown in \autoref{fig:appen-zeroshotstat}, the HICO-DET (UF-RC) scenario follows an imbalanced distribution, resulting in high variance. In contrast, \ours~demonstrates a lower variance compared to HICO-DET.

\paragraph{Why GEN-VLKT Underperforms Compared to HOICLIP Despite Both Using CLIP?}
Both HOICLIP and GEN-VLKT integrate CLIP-based knowledge to enhance their zero-shot generalization capabilities. However, their performances in our \ours~balanced zero-shot test set diverge significantly, with HOICLIP achieving a competitive mAP of 28.36, while GEN-VLKT records a drastic decline to 4.38 mAP. This difference can be attributed to several key differences in their architectural and training strategies.

Firstly, GEN-VLKT uses a global feature distillation process that tends to overfit the model to the specific interactions present in the train data. This overfitting limits its ability to generalize to novel verb–object combinations required in a balanced zero-shot test. In contrast, HOICLIP maintains a consistent classifier dimension setup for all 600 HOI classes from the outset, allowing it to actively predict across the entire class space, including unseen classes. This consistent setup enables HOICLIP to better handle the uniform distribution of unseen classes in the balanced zero-shot test.

Secondly, GEN-VLKT struggles with capturing localized interaction cues essential for distinguishing unseen combinations. While CLIP is good at aligning global image-text embeddings, GEN-VLKT's architecture does not sufficiently incorporate mechanisms to model fine-grained, compositional interactions. This limitation seems to be exacerbated by its classifier expansion strategy, where classifiers are frozen to match the dimensions of seen classes during train and only expanded at test. As a result, GEN-VLKT fails to effectively utilize the expanded classifiers for unseen classes, leading to a significant coverage gap. On the other hand, HOICLIP leverages its architecture to maintain robust interaction representations that generalize more effectively to unseen classes without relying solely on global features. This architectural advantage allows HOICLIP to better manage the balanced zero-shot set, where the absence of class frequency biases necessitates a more uniform generalization approach.

In summary, the underperformance of GEN-VLKT compared to HOICLIP in the balanced zero-shot set highlights the importance of architectural choices that facilitate both global and localized feature integration, as well as training strategies that prevent overfitting to seen interactions. These factors are crucial for achieving robust generalization across a uniformly distributed set of unseen verb–object pairs.

\section{Top-K HOI categories in \ours~ dataset}
In \autoref{tab:hoi_category}, we provide the Top-351 HOI class list for balancing process. The HOI class list consists of 351 HOI class IDs and their corresponding (verb, object) pairs, where the HOI class IDs are provided at \href{https://umich-ywchao-hico.github.io/}{official HICO-DET dataset websites}.

\begin{table*}[]
\centering
\tiny
\caption{Top-351 HOI class IDs and (verb, object) table in \ours~dataset.}
\begin{tabular}{cc|cc|cc|cc}
ID  &(verb, object) & ID  &(verb, object) & ID  &(verb, object) & ID&  (verb, object) \\ \hline
1	&(board, airplane)&	141	&(run, horse)&	288	&(talk\_on, cell phone)&	443	&(hold, sandwich)\\
2	&(direct, airplane)&	142	&(straddle, horse)&	289	&(text\_on, cell phone)&	445	&(no\_interaction, sandwich)\\
3	&(exit, airplane)&	143	&(train, horse)&	291	&(check, clock)&	446	&(cut\_with, scissors)\\
4	&(fly, airplane)&	144	&(walk, horse)&	292	&(hold, clock)&	447	&(hold, scissors)\\
5	&(inspect, airplane)&	146	&(no\_interaction, horse)&	295	&(no\_interaction, clock)&	448	&(open, scissors)\\
6	&(load, airplane)&	147	&(hold, motorcycle)&	296	&(carry, cup)&	453	&(no\_interaction, sink)\\
7	&(ride, airplane)&	148	&(inspect, motorcycle)&	297	&(drink\_with, cup)&	454	&(carry, skateboard)\\
8	&(sit\_on, airplane)&	149	&(jump, motorcycle)&	298	&(hold, cup)&	455	&(flip, skateboard)\\
10	&(no\_interaction, airplane)&	151	&(park, motorcycle)&	301	&(sip, cup)&	456	&(grind, skateboard)\\
12	&(hold, bicycle)&	153	&(race, motorcycle)&	303	&(fill, cup)&	457	&(hold, skateboard)\\
13	&(inspect, bicycle)&	154	&(ride, motorcycle)&	305	&(no\_interaction, cup)&	458	&(jump, skateboard)\\
14	&(jump, bicycle)&	155	&(sit\_on, motorcycle)&	307	&(carry, donut)&	459	&(pick\_up, skateboard)\\
17	&(push, bicycle)&	156	&(straddle, motorcycle)&	308	&(eat, donut)&	460	&(ride, skateboard)\\
18	&(repair, bicycle)&	157	&(turn, motorcycle)&	309	&(hold, donut)&	462	&(stand\_on, skateboard)\\
19	&(ride, bicycle)&	160	&(no\_interaction, motorcycle)&	310	&(make, donut)&	463	&(no\_interaction, skateboard)\\
20	&(sit\_on, bicycle)&	161	&(carry, person)&	311	&(pick\_up, donut)&	465	&(carry, skis)\\
21	&(straddle, bicycle)&	163	&(hold, person)&	313	&(no\_interaction, donut)&	466	&(hold, skis)\\
22	&(walk, bicycle)&	164	&(hug, person)&	314	&(feed, elephant)&	468	&(jump, skis)\\
24	&(no\_interaction, bicycle)&	165	&(kiss, person)&	315	&(hold, elephant)&	471	&(ride, skis)\\
25	&(chase, bird)&	168	&(teach, person)&	320	&(pet, elephant)&	472	&(stand\_on, skis)\\
26	&(feed, bird)&	170	&(no\_interaction, person)&	321	&(ride, elephant)&	473	&(wear, skis)\\
27	&(hold, bird)&	174	&(no\_interaction, potted plant)&	322	&(walk, elephant)&	476	&(carry, snowboard)\\
30	&(watch, bird)&	176	&(feed, sheep)&	323	&(wash, elephant)&	477	&(grind, snowboard)\\
31	&(no\_interaction, bird)&	177	&(herd, sheep)&	324	&(watch, elephant)&	478	&(hold, snowboard)\\
32	&(board, boat)&	178	&(hold, sheep)&	330	&(no\_interaction, fire hydrant)&	479	&(jump, snowboard)\\
33	&(drive, boat)&	183	&(shear, sheep)&	331	&(hold, fork)&	480	&(ride, snowboard)\\
35	&(inspect, boat)&	184	&(walk, sheep)&	332	&(lift, fork)&	481	&(stand\_on, snowboard)\\
37	&(launch, boat)&	186	&(no\_interaction, sheep)&	336	&(no\_interaction, fork)&	482	&(wear, snowboard)\\
38	&(repair, boat)&	187	&(board, train)&	337	&(block, frisbee)&	484	&(hold, spoon)\\
39	&(ride, boat)&	188	&(drive, train)&	338	&(catch, frisbee)&	485	&(lick, spoon)\\
40	&(row, boat)&	191	&(ride, train)&	339	&(hold, frisbee)&	487	&(sip, spoon)\\
41	&(sail, boat)&	192	&(sit\_on, train)&	341	&(throw, frisbee)&	488	&(no\_interaction, spoon)\\
42	&(sit\_on, boat)&	194	&(no\_interaction, train)&	342	&(no\_interaction, frisbee)&	489	&(block, sports ball)\\
43	&(stand\_on, boat)&	197	&(watch, tv)&	343	&(feed, giraffe)&	490	&(carry, sports ball)\\
44	&(tie, boat)&	198	&(no\_interaction, tv)&	347	&(watch, giraffe)&	491	&(catch, sports ball)\\
46	&(no\_interaction, boat)&	201	&(eat, apple)&	348	&(no\_interaction, giraffe)&	492	&(dribble, sports ball)\\
47	&(carry, bottle)&	202	&(hold, apple)&	349	&(hold, hair drier)&	493	&(hit, sports ball)\\
48	&(drink\_with, bottle)&	203	&(inspect, apple)&	350	&(operate, hair drier)&	494	&(hold, sports ball)\\
49	&(hold, bottle)&	208	&(no\_interaction, apple)&	353	&(carry, handbag)&	495	&(inspect, sports ball)\\
50	&(inspect, bottle)&	209	&(carry, backpack)&	354	&(hold, handbag)&	496	&(kick, sports ball)\\
54	&(no\_interaction, bottle)&	210	&(hold, backpack)&	357	&(carry, hot dog)&	498	&(serve, sports ball)\\
55	&(board, bus)&	213	&(wear, backpack)&	358	&(cook, hot dog)&	501	&(throw, sports ball)\\
57	&(drive, bus)&	214	&(no\_interaction, backpack)&	360	&(eat, hot dog)&	502	&(no\_interaction, sports ball)\\
60	&(load, bus)&	215	&(buy, banana)&	361	&(hold, hot dog)&	506	&(no\_interaction, stop sign)\\
61	&(ride, bus)&	216	&(carry, banana)&	362	&(make, hot dog)&	507	&(carry, suitcase)\\
62	&(sit\_on, bus)&	218	&(eat, banana)&	363	&(no\_interaction, hot dog)&	508	&(drag, suitcase)\\
65	&(no\_interaction, bus)&	219	&(hold, banana)&	364	&(carry, keyboard)&	509	&(hold, suitcase)\\
68	&(drive, car)&	220	&(inspect, banana)&	366	&(hold, keyboard)&	514	&(pick\_up, suitcase)\\
69	&(hose, car)&	224	&(no\_interaction, banana)&	367	&(type\_on, keyboard)&	516	&(no\_interaction, suitcase)\\
70	&(inspect, car)&	225	&(break, baseball bat)&	368	&(no\_interaction, keyboard)&	517	&(carry, surfboard)\\
74	&(ride, car)&	226	&(carry, baseball bat)&	369	&(assemble, kite)&	519	&(hold, surfboard)\\
75	&(wash, car)&	227	&(hold, baseball bat)&	370	&(carry, kite)&	524	&(ride, surfboard)\\
76	&(no\_interaction, car)&	229	&(swing, baseball bat)&	371	&(fly, kite)&	525	&(stand\_on, surfboard)\\
79	&(hold, cat)&	231	&(wield, baseball bat)&	372	&(hold, kite)&	528	&(no\_interaction, surfboard)\\
80	&(hug, cat)&	232	&(no\_interaction, baseball bat)&	373	&(inspect, kite)&	529	&(carry, teddy bear)\\
82	&(pet, cat)&	233	&(hold, baseball glove)&	374	&(launch, kite)&	530	&(hold, teddy bear)\\
83	&(scratch, cat)&	234	&(wear, baseball glove)&	375	&(pull, kite)&	531	&(hug, teddy bear)\\
86	&(no\_interaction, cat)&	237	&(hunt, bear)&	376	&(no\_interaction, kite)&	533	&(no\_interaction, teddy bear)\\
87	&(carry, chair)&	238	&(watch, bear)&	377	&(cut\_with, knife)&	534	&(carry, tennis racket)\\
88	&(hold, chair)&	241	&(lie\_on, bed)&	378	&(hold, knife)&	535	&(hold, tennis racket)\\
89	&(lie\_on, chair)&	242	&(sit\_on, bed)&	379	&(stick, knife)&	537	&(swing, tennis racket)\\
90	&(sit\_on, chair)&	245	&(lie\_on, bench)&	381	&(wield, knife)&	538	&(no\_interaction, tennis racket)\\
92	&(no\_interaction, chair)&	246	&(sit\_on, bench)&	383	&(no\_interaction, knife)&	539	&(adjust, tie)\\
94	&(lie\_on, couch)&	247	&(no\_interaction, bench)&	384	&(hold, laptop)&	541	&(hold, tie)\\
95	&(sit\_on, couch)&	248	&(carry, book)&	385	&(open, laptop)&	545	&(wear, tie)\\
96	&(no\_interaction, couch)&	249	&(hold, book)&	386	&(read, laptop)&	546	&(no\_interaction, tie)\\
98	&(herd, cow)&	250	&(open, book)&	388	&(type\_on, laptop)&	555	&(sit\_on, toilet)\\
99	&(hold, cow)&	251	&(read, book)&	389	&(no\_interaction, laptop)&	558	&(no\_interaction, toilet)\\
102	&(lasso, cow)&	252	&(no\_interaction, book)&	394	&(control, mouse)&	559	&(brush\_with, toothbrush)\\
104	&(pet, cow)&	253	&(hold, bowl)&	395	&(hold, mouse)&	560	&(hold, toothbrush)\\
106	&(walk, cow)&	257	&(no\_interaction, bowl)&	397	&(no\_interaction, mouse)&	567	&(no\_interaction, traffic light)\\
107	&(no\_interaction, cow)&	259	&(eat, broccoli)&	401	&(hold, orange)&	569	&(drive, truck)\\
109	&(eat\_at, dining table)&	260	&(hold, broccoli)&	407	&(no\_interaction, orange)&	570	&(inspect, truck)\\
110	&(sit\_at, dining table)&	265	&(blow, cake)&	410	&(inspect, oven)&	571	&(load, truck)\\
111	&(no\_interaction, dining table)&	266	&(carry, cake)&	414	&(no\_interaction, oven)&	573	&(ride, truck)\\
112	&(carry, dog)&	267	&(cut, cake)&	418	&(no\_interaction, parking meter)&	574	&(sit\_on, truck)\\
116	&(hold, dog)&	268	&(eat, cake)&	420	&(carry, pizza)&	576	&(no\_interaction, truck)\\
118	&(hug, dog)&	269	&(hold, cake)&	421	&(cook, pizza)&	577	&(carry, umbrella)\\
121	&(pet, dog)&	272	&(pick\_up, cake)&	422	&(cut, pizza)&	578	&(hold, umbrella)\\
123	&(scratch, dog)&	273	&(no\_interaction, cake)&	423	&(eat, pizza)&	583	&(stand\_under, umbrella)\\
125	&(train, dog)&	274	&(carry, carrot)&	424	&(hold, pizza)&	584	&(no\_interaction, umbrella)\\
126	&(walk, dog)&	275	&(cook, carrot)&	425	&(make, pizza)&	588	&(no\_interaction, vase)\\
129	&(no\_interaction, dog)&	277	&(eat, carrot)&	426	&(pick\_up, pizza)&	590	&(hold, wine glass)\\
132	&(hold, horse)&	278	&(hold, carrot)&	429	&(no\_interaction, pizza)&	591	&(sip, wine glass)\\
134	&(jump, horse)&	283	&(no\_interaction, carrot)&	431	&(hold, refrigerator)&	592	&(toast, wine glass)\\
138	&(pet, horse)&	284	&(carry, cell phone)&	435	&(hold, remote)&	595	&(no\_interaction, wine glass)\\
139	&(race, horse)&	285	&(hold, cell phone)&	436	&(point, remote)&	599	&(watch, zebra)\\
140	&(ride, horse)&	286	&(read, cell phone)&	442	&(eat, sandwich)&

\end{tabular}

\label{tab:hoi_category}
\end{table*}


\end{document}